\documentclass[11pt]{article}

\usepackage{epsfig,amsmath,latexsym,amssymb}
\usepackage{graphicx}
\usepackage{lscape}
\usepackage{picture, eso-pic, tikz} 
\usepackage{dsfont}
\usepackage{appendix}
\usepackage{multirow}
\usepackage{listings}
\usepackage{adjustbox}
\usepackage{rotating}
\usepackage{apacite}
\usepackage{longtable}
\usepackage{booktabs}
\usepackage{colortbl}
\usepackage{soul}
\usepackage{booktabs}
\usepackage{caption}
\usepackage{longtable}
\usepackage{afterpage}

\usepackage{epsfig,amsmath,latexsym,amssymb}
\usepackage{graphicx}
\usepackage{lscape}
\usepackage{picture, eso-pic, tikz} 
\usepackage{dsfont}
\usepackage{appendix}
\usepackage{multirow}
\usepackage{apacite}
\usepackage{listings}
\usepackage{adjustbox}
\usepackage{amsmath}
\usepackage{bbm}
\usepackage{float}
\usepackage{multirow}
\usepackage{rotating}
\usepackage{booktabs}
\usepackage{colortbl}
\usepackage{soul}
\usepackage{booktabs}
\usepackage{caption}
\usepackage{longtable}

\usepackage{etoolbox}
\BeforeBeginEnvironment{appendices}{\clearpage}

\let\cite\shortcite
\let\citeA\shortciteA
\AtBeginDocument{}
\RequirePackage[hyphens]{url} 

\usepackage{etoolbox}

\AtBeginEnvironment{APACrefURL}{\renewcommand{\url}[1]{}}

\AtBeginEnvironment{APACrefDOI}{\renewcommand{\doi}[1]{}}

\usepackage{etoolbox}
\BeforeBeginEnvironment{appendices}{\clearpage}

\usepackage{relsize}
\newcommand*{\defeq}{\stackrel{\mathsmaller{\mathsf{def}}}{=}}
\newcommand*{\seteq}{\stackrel{\mathsmaller{\mathsf{set}}}{=}}

\def\R{{\mathbb R}}  
\def\N{{\mathbb N}}  
\def\E{{\mathbb E}}  %

\newcommand{\Remm}[1]{}
\newtheorem{theo}{Theorem}[section]

\newtheorem{model ass}[theo]{Model Assumptions}
\newtheorem{ass}[theo]{Assumptions}

\newtheorem{rems}[theo]{Remark}

\numberwithin{equation}{section}

\definecolor{MyGray}{rgb}{0.92,0.92,0.92}


\def\bx{\boldsymbol{x}}
\def\be{\boldsymbol{e}}
\def\by{\boldsymbol{y}}
\def\bz{\boldsymbol{z}}

\begin{document}

	\author{
Ronald Richman\footnote{Old Mutual Insure and University of the Witwatersrand, Johannesburg, South Africa; ronaldrichman@gmail.com}
	\and
	Mario V.~W\"uthrich\footnote{RiskLab, Department of Mathematics, ETH Zurich, Switzerland;
		mario.wuethrich@math.ethz.ch}
}

\date{Version of \today}
\title{Smoothness and monotonicity constraints for neural networks using ICEnet}
\maketitle

\begin{abstract}
	\noindent	
Deep neural networks have become an important tool for use in actuarial tasks, due to the significant gains in accuracy provided by these techniques compared to traditional methods, but also due to the close connection of these models to the Generalized Linear Models (GLMs) currently used in industry. Whereas constraining GLM parameters relating to insurance risk factors to be smooth or exhibit monotonicity is trivial, methods to incorporate such constraints into deep neural networks have not yet been developed. This is a barrier for the adoption of neural networks in insurance practice since actuaries often impose these constraints for commercial or statistical reasons. In this work, we present a novel method for enforcing constraints within deep neural network models, and we show how these models can be trained. Moreover, we provide example applications using real-world datasets. We call our proposed method \textit{ICEnet} to emphasize the close link of our proposal to the individual conditional expectation (ICE) model interpretability technique.
	
\medskip

\noindent
{\bf Keywords.} Smoothing, Whittaker--Henderson Smoothing, Graduation, Monotonicity, Deep Neural Networks, Constrained Likelihood, Individual Conditional Expectation 
	
\end{abstract}

\section{Introduction}
Deep neural networks have recently emerged as a promising technique for use in tasks across the various traditional disciplines of actuarial science, including pricing, reserving, experience analysis and mortality forecasting. Moreover, deep learning has been applied in emerging areas of actuarial practice, such as analysis of telematics data, natural language processing and image recognition. These techniques provide significant gains in accuracy compared to traditional methods, while the close connection of these models to the Generalized Linear Models (GLMs) currently used in industry enable easier understanding of these models for classically trained actuaries than other machine learning paradigms such as boosted trees. 

On the other hand, one seeming disadvantage of neural network models is that the output from these models may exhibit undesirable characteristics for actuarial purposes. A first issue is that predictions may vary in a rough manner with changes in insurance risk factors. In some contexts, such as general insurance pricing, this may be problematic to explain to customers, intermediaries or other stakeholders, or may indicate problems with data credibility. For example, consider a general insurance pricing model that uses driver age as a risk factor. Usually, from a commercial perspective, as a customer ages, it is expected that her insurance rates would change smoothly, however, unconstrained output from neural networks may produce rates that vary roughly with age. It is difficult to explain to customers why rates might increase one year, then decrease the next, and, moreover, extra costs might arise from needing to address these types of queries. A second issue is that actuaries often wish to constrain predictions from models to increase (or decrease) in a monotonic manner with some risk factors, for example, increasing sums insured should lead, other things being equal, to higher average costs per claim and worsening bonus-malus scores should imply higher expected frequencies of claims. Whereas constraining GLM parameters relating to insurance risk factors to be smooth or exhibit monotonicity is trivial, since the coefficients of GLMs can be modified directly, methods to introduce these constraints into deep neural networks have not yet been developed. Thus, a significant barrier for the adoption of neural networks in practice exists.

Typical practice when manually building actuarial models is that actuaries will attempt to enforce smoothness or monotonicity constraints within models by modifying model structure or coefficients manually, as mentioned before, by applying relevant functional forms within models, such as parametric functions, regularization or by applying post-hoc smoothing techniques; we mention, e.g., Whittaker--Henderson smoothing  \cite{Whittaker, Henderson} which adds regularization to parameters. In the case of GLMs for general insurance pricing, the simple linear structure of these models is often exploited by, firstly, fitting an unconstrained GLM, then specifying some simplifications whereby coefficients are forced to follow the desired functional form or meet a smoothness criteria. In more complex models, such as neural networks, operating directly on model coefficients or structure is less feasible.

To overcome this challenge, here we present methods for enforcing smoothness and monotonicity constraints within deep neural network models. The key idea can be summarized as follows: as a first step insurance risk factors that should be constrained are identified, then, the datasets used for model fitting are augmented to produce pseudo-data that reflect the structure of the identified variables. In the next step, we design a multi-input/multi-output neural network that can process jointly the original observation as well as the pseudo-data. Finally, a joint loss function is used to train the network to make accurate predictions while enforcing the desired constraints on the pseudo-data. We show how these models can be trained and provide example applications using a real-world general insurance pricing dataset.  

The method we propose can be related to the Individual Conditional Expectation (ICE) model interpretability technique of \citeA{goldstein2015peeking}, therefore, we call our proposal the \textit{ICEnet}, to emphasize this connection.  To enable the training of neural networks with constraints, the ICEnet structures networks to output a vector of predictions derived from the pseudo-data input to the network; these predictions derived from the network for the key variables of interest on the pseudo-data are exactly equivalent to the outputs used to derive an ICE plot. The selected constraints then constrain these ICEnet outputs from the network to be as smooth or monotonic as required. For this purpose, we will use Fully Connected Networks (FCNs) with embedding layers for categorical variables to perform supervised learning, and, in the process, create the ICEnet outputs for the variables of interest.

\bigskip

\textbf{Literature review.} Practicing actuaries have often smoothed results of experience analyses in both general and life insurance, usually on the assumption that outputs of models that do not exhibit smoothness are anomalous. For example, \citeA{goldburd2016generalized} and \citeA{anderson2007practitioner} discuss manual smoothing methods in the context of general insurance pricing with GLMs. In life insurance, two main types of techniques are used to graduate (i.e. smooth) mortality tables; in some jurisdictions, such as the United Kingdom and South Africa, combinations of polynomial functions have often been used to fit mortality data directly, see, for example, \citeA{Forfar1987}, whereas post-hoc smoothing methods such as Whittaker--Henderson smoothing  \cite{Whittaker, Henderson} have often been used in the United States. The use of splines and Generalized Additive Models (GAMs) have also been considered for these purposes, see \citeA{goldburd2016generalized} in the context of general insurance and \citeA{Debon2006} in the context of mortality data. 

More recently, penalized regression techniques have been utilized for variable selection and categorical level fusion for general insurance pricing. These are often based on the Least Absolute Shrinkage and Selection Operator (LASSO) regression of \citeA{tibslasso} and its extensions \cite{lasso}; here we mention the fused LASSO which produces model coefficients that vary smoothly, see \citeA{tibshirani2005sparsity}, which is particularly useful for deriving smoothly varying models for ordinal categorical data. In the actuarial literature, example applications of the constrained regression approach are \citeA{DEVRIENDT2021248}, who propose an algorithm for incorporating multiple penalties for complex general insurance pricing data and \citeA{Antonio}, who provide a data-driven binning approach designed to produce GLMs which closely mirror smooth GAMs. 

Within the machine and deep learning literature, various methods for enforcing monotonicity constraints within Gradient Boosting Machines (GBM) and neural network models have been proposed. Monotonicity constraints are usually added to GBMs by modifying the process used to fit decision trees when producing these models; an example of this can be found in the well-known XGBoost library of \citeA{chen2016xgboost}. Since these constraints are implemented specifically by modifying how the decision trees underlying the GBM are fit to the data, the same process cannot be applied for neural network models. Within the deep learning literature, monotonicity constraints have been addressed by constraining the weights of neural networks to be positive, see, for example, \citeA{sill1997monotonic} and \citeA{Daniels}, who generalize the earlier methods of \citeA{sill1997monotonic}, or by using specially designed networks, such as the lattice networks of \citeA{you2017deep}. In the finance literature, \citeA{Kellner} proposes a method to ensure monotonicity of multiple quantiles output from a neural network, by adding a monotonicity constraint to multiple outputs from the network; this idea is similar to what we implement below. Another approach to enforcing monotonicity involves post-processing the outputs of machine learning models with a different algorithm, such as isotonic regression; see \citeA{wuthrich2023isotonic} for a recent application of this to ensure that outputs of general insurance pricing models are autocalibrated. 

On the other hand, the machine and deep learning literature seemingly has not addressed the issue of ensuring that predictions made with these models vary smoothly, and, moreover, the methods proposed within the literature for monotonicity constraints cannot be directly applied to enforce smoothness constraints. Thus, the ICEnet proposal of this work, which adds flexible penalties to a specially designed neural network, fills a gap in the literature by showing how both monotonicity and smoothness constraints can be enforced with the same method in deep learning models.

\bigskip

\textbf{Structure of the manuscript.} The rest of the manuscript is structured as follows. Section \ref{Methodology} provides notation and discusses neural networks and machine learning interpretability. Section \ref{ICEnet_section} defines the ICEnet, which is applied to the French Motor Third Party Liability data in Section \ref{applying}. A local approximation to the ICEnet is presented in Section \ref{Local}. Discussion of the results and conclusions are given in Section \ref{conclusions}. The supplementary provides the code and further numerical analysis of the ICEnet.

\section{Neural networks and Individual Conditional Expectations} \label{Methodology}
We begin by briefly introducing supervised learning with neural networks, expand these definitions to FCNs using embedding layers and discuss their training process. With these building blocks, we then present the ICEnet proposal in the next section.
 
\subsection{Supervised learning and neural networks} \label{supervised}

We work in the usual setup of supervised learning. Independent observations $y_n \in \R$ of a variable of interest have been made for instances (insurance policies) $1 \le n \le N$. In addition, covariates $\bx_n$ have been collected for all instances  $1\le n \le N$. These covariates can be used to create predictions $\widehat{y}_n \in \R$ of the variable of interest. In what follows, note that we drop the subscript from $\bx_n$ for notational convenience. The covariates in $\bx$ are usually of two main types in actuarial tasks: the first of these are real-valued covariates $\bx^{[r]} \in \R^{q_r}$, where the superscript $[\cdot]$ represents the subset of the vector $\bx$, $r$ is the set of real-valued covariates and where there are $q_r$ real-valued variables. The second type of covariates are categorical, which we assume have been coded as positive integers; we represent these as $\bx^{[c]} \in \N^{q_c}$, where the set of categorical covariates is $c$. Thus, $\bx = (\bx^{[r]}, \bx^{[c]})$. In actuarial applications, often predictions are proportional to a scalar unit of exposure $v_n \in \R^+$, thus, each observation is a tensor $(y, \bx^{[r]}, \bx^{[c]}, v) \in \R \times \R^{q_r} \times \N^{q_c} \times \R^+$.

In this work, we will use deep neural networks, which are efficient function approximators, for predicting $y$. We represent the general class of neural networks as $\Psi_W(\bx)$, where $W$ are the network parameters (weights and biases). Using these, we aim to study the
regression function
\begin{equation*}
 \Psi_W: \R^{q_r} \times \N^{q_c}   \to \R, \qquad \bx ~\mapsto~ \widehat{y} = \Psi_W(\bx)\,v.
\end{equation*}

We follow Chapter 7 of \citeA{Wuthrich2021} for the notation defining neural networks. Neural networks are machine learning models constructed by composing non-linear functions (called layers) operating on the vector $\bx$, which is the input to the network. A network consisting of only a single layer of non-linear functions has depth $d=1$, and is called a shallow network. More complex networks consisting of multiple layers with depth $d \ge 2$ are called deep networks. We denote the $i$-th layer by $\bz^{(i)}$. These non-linear functions (layers) transform the input variables $\bx$ into new representations, which are optimized to perform well on the supervised learning task, using a process called representation learning. Representation learning can be denoted as the composition
\begin{equation*}
\bx~\mapsto ~   \bz^{(d:1)}(\bx)~\defeq~\left(\bz^{(d)} \circ \dots \circ \bz^{(1)} \right)(\bx) ~\in~ \R^{q_d},
\end{equation*}
where $d \in \N$ is the number of layers $\bz^{(i)}$ of the network and $q_i \in \N$ are the dimensions of these layers for $1\le i \le d$. Thus, each layer $\bz^{(i)}:\R^{q_{i-1}} \to \R^{q_i}$ transforms the representation at the previous stage to a new, modified representation. 

FCNs define the $j$-th component (called neuron or unit) of each layer $\bz^{(i)}$, $1\le j \le q_i$, as the mapping
\begin{equation}\label{FCN_neurons}
  \bz=(z_1,\ldots, z_{q_{i-1}})^\top \in \R^{q_{i-1}}
  \quad \mapsto \quad  z^{(i)}_j(\bz) = \phi \left (\sum_{k = 1}^{q_{i-1}} w^{(i)}_{j,k} z_k + b^{(i)}_j \right),
\end{equation}
for a non-linear activation function $\phi:\R\to\R$, and where $z^{(i)}_j(\cdot)$ is the $j$-th component of layer $\bz^{(i)}(\cdot)$, $w^{(i)}_{j,k}\in\R$ is the regression weight of this $j$-th neuron connecting to the $k$-th neuron of the previous layer, $z_k=z^{(i-1)}_k$, and $b^{(i)}_j\in \R$ is the intercept or bias for the $j$-th neuron in layer $\bz^{(i)}$. It can be seen from \eqref{FCN_neurons} that the neurons $z^{(i)}_j(\cdot)$ of a FCN  connect to all of the neurons $z^{(i-1)}_k(\cdot)$ in the previous layer $\bz^{(i-1)}$ through the weights $w^{(i)}_{j,k}$, explaining the description of these networks as ``fully-connected".

Combining these layers, a generic FCN regression function can be defined as follows
\begin{equation}\label{FFN generic}
  \bx ~ \mapsto ~
   \Psi_W(\bx) ~\defeq~  g^{-1} \left( \sum_{k=1}^{q_d} w^{(d+1)}_k z_k^{(d:1)}(\bx) + b^{(d+1)}\right),
 \end{equation} where $g^{-1}(\cdot)$ is a suitably chosen inverse link function that transforms the outputs of the network to the scale of the observations $y$. The notation $W$ in $\Psi_W(\bx)$ indicates that we collect all the weights $w^{(i)}_{j,k}$ and biases $ b^{(i)}_j$ in $W$, giving us a network parameter of dimension $(q_{d}+1) + \sum_{i=1}^d (q_{i-1}+1)q_i$, where $q_0$ is the dimension of the input $\bx$.

For most supervised learning applications in actuarial work, a neural network of the form \eqref{FCN_neurons}-\eqref{FFN generic} is applied to the covariate $\bx$ to create a single prediction $\widehat{y} = \Psi_W(\bx)v$. Below, we will define how the same network $\Psi_W(\cdot)$ can be applied to a vector of observations to produce multiple predictions, which will be used to constrain the network.

State-of-the-art neural network calibration is performed using a Stochastic Gradient Descent (SGD) algorithm, performed on mini-batches of observations. To calibrate the network, an appropriate loss function $L(\cdot, \cdot)$ must be selected. For general insurance pricing, the loss function is often the deviance loss function of an exponential-family distribution, such as the Poisson distribution for frequency modelling or the Gamma distribution for severity modelling. For more details on the SGD procedure, we refer to \citeA{Goodfellow}, and for a detailed explanation of exponential-family modelling for actuarial purposes see Chapters 2 and 4 of \citeA{Wuthrich2021}.

\subsection{Pre-processing covariates for FCNs} \label{processing}
For the following, we assume that the $N$ instances of $\bx_n$, $1\le n \le N$, have been collected into a matrix $X = [X^{[r]},X^{[c]}] \in \R^{N \times (q_r+q_c)}$, where $X^{[\cdot]}$ represents a subset of the columns of $X$. Thus, to select the $j$-th column of $X$, we write $X^{[j]}$ and, furthermore, to represent the $n$-th row of $X$, we will write $X_n = \bx_n$. We also assume that all of the observations of $y_n$ have been collected into a vector $\by=(y_1,\ldots, y_N)^\top \in \R^N$.

\subsubsection{Categorical covariates}

The types of categorical data comprising $X^{[c]}$ are usually either qualitative (nominal) data with no inherent ordering, such as type of motor vehicle, or ordinal data with an inherent ordering, such as bad-average-good driver. Different methods for pre-processing categorical data for use within machine learning methods have been developed, with one-hot encoding being a popular choice for traditional machine learning methods. Here, we focus on the categorical embedding technique (entity embedding) of \citeA{guo2016entity}; see \citeA{richman2021ai1} and \citeA{Delong} for a brief overview of other options and an introduction to embeddings. Assume that the $t$-th categorical variable, corresponding to column $X^{[c_t]}$, for $ 1 \leq t \leq q_c$, can take one of $K_t \geq 2$ values in the set of levels $\{a^t_1, \dots, a^t_{K_t} \}$. An embedding layer for this categorical variable maps each member of the set to a low-dimensional vector representation of dimension $b_t < K_t$, i.e., 
\begin{equation}\label{embedding layer}
 \be: \{a^t_1, \dots, a^t_{K_t} \}   \to \R^{b_t}, \qquad a^t_k ~\mapsto~ \be(a^t_k) \defeq \be^{t(k)},
\end{equation}
meaning to say that the $k$-th level $a^t_k$ of the $t$-th categorical variable receives a low dimensional vector representation $\be^{t(k)}\in \R^{b_t}$. When utilizing an embedding layer within a neural network, we calibrate the $K_tb_t$ parameters of the embedding layer as part of fitting the weights $W$ of the network $\Psi_W(\bx)$. Practically, when inputting a categorical covariate to a neural network, each level of the covariate is mapped to a unique natural number, thus, we have represented these covariates as $\bx^{[c]} \in \N^{q_c}$, and these represenations are then embedded according \eqref{embedding layer} which can be interpreted as an additional layer of the network; this is graphically illustrated in Figure 7.9 of \citeA{Wuthrich2021}.

\subsubsection{Numerical covariates}

To enable the easy calibration of neural networks, numerical covariates must be scaled to be of similar magnitudes. In what follows, we will use the min-max normalization, where each raw numerical covariate $\dot{X}^{[r_t]}$ is scaled to 
\begin{equation*}
X^{[r_t]} = \frac{\dot{X}^{[r_t]} - \min(\dot{X}^{[r_t]})}{\max(\dot{X}^{[r_t]}) - \min(\dot{X}^{[r_t]})},   
\end{equation*}
where $\dot{X}^{[r_t]}$ is $t$-th raw (i.e., unscaled) continuous covariate and the operation is performed for each element of column $\dot{X}^{[r_t]}$ of the matrix of raw continuous covariates, $\dot{X}^{[r]}$.  

We note that the continuous data comprising $\dot{X}^{[r]}$ can also easily be converted to ordinal data through binning, for example, by mapping each observed covariate in the data to one of the quantiles of that covariate. We do not consider this option for processing continuous variables in this work, however, we will use quantile binning to discretize the continuous covariates $X^{[r]}$ for the purpose of estimating the ICEnets used here.

\subsection{Individual Conditional Expectations and Partial Dependence Plots} \label{ICE_method}

Machine learning interpretability methods are used to explain how a machine learning model, such as a neural network, has estimated the relationship between the covariates $\bx$ and the predictions $\widehat{y}$; for an overview of these, see \citeA{biecek2021explanatory}. Two related methods for interpreting machine learning models are the Partial Dependence Plot (PDP) of \citeA{Friedman2001} and the Individual Conditional Expectations (ICE) method of \citeA{goldstein2015peeking}. The ICE method estimates how the predictions $\widehat{y}_n$ for each instance $1 \leq n \leq N$ change as a single component of the covariates $\bx_n$ is varied over its observed range of possible values, while holding all of the other components of $\bx_n$ constant. The PDP method is simply the average over all of the individual ICE outputs of the instances $1\le n \le N$ derived in the previous step. By inspecting the resultant ICE and PDP outputs, the relationship of the predictions with a single component of $\bx$ can be observed; by performing the same process for each component of $\bx$, the relationships with all of the covariates can be shown.

To estimate the ICE for a neural network, we now consider the creation of pseudo-data that will be used to create the ICE output of the network. We consider one column of $X$, denoted as $X^{[j]}$, $1\le j \le q_r+q_c$. If the selected column $X^{[j]}$ is categorical, then, as above, the range of values which can be taken by each element of $X^{[j]}$ are simply the values in the set of levels $\{a^j_1, \dots, a^j_{K_j} \}$. If the selected column $X^{[j]}$ is continuous, then we assume that a range of values for the column has been selected using quantile binning or another procedure, and we denote the set of these values using the same notation, i.e., $\{a^j_1, \dots, a^j_{K_j} \}$, where $K_j\in \N$ is
the number of selected values $a_u^j \in \R$ and where we typically assume that the continuous variables in $\{a^j_1, \dots, a^j_{K_j} \}$
are ordered in increasing order. 

We define $\Tilde{X}^{[j]}(u)$ as a copy of $X$, where all of the components of $X$ remain the same, except for the $j$-th column of $X$, which is set to the $u$-th value of the set $\{a^j_1, \dots, a^j_{K_j} \}$, $1 \leq u \leq K_j$. By sequentially creating predictions using a (calibrated) neural network applied to the pseudo-data, $\Psi_W(\Tilde{X}^{[j]}(u))$ (where the network is applied in a row-wise manner), we are able to derive the ICE outputs for each variable of interest, as we vary the value of $1\le u \le K_j$. In particular, by allowing $a_u^j$ to take each value in the set  $\{a^j_1, \dots, a^j_{K_j} \}$, for each instance $1\le n \le N$ separately, we define the ICE for covariate $j$ as a vector of predictions on the artificial data, $\widetilde{\by}^{[j]}_n$, as
\begin{equation}\label{ICE_individual_n}
 \widetilde{\by}^{[j]}_n = \left[\Psi_W(\Tilde{\bx}_n^{[j]}(1))v_n, \ldots, \Psi_W(\Tilde{\bx}_n^{[j]}(K_j))v_n  \right],
\end{equation} 
where $\Tilde{\bx}_n^{[j]}(u)$ represents the vector of covariates for instance $n$, where the $j$-th entry has been set equal to the $u$-th value $a_u^j$ of that covariate, which is contained in the appropriate set $\{a^j_1, \dots, a^j_{K_j} \}$.

The PDP can then be derived from \eqref{ICE_individual_n} by averaging the ICE outputs over all instances. We set
\begin{equation}\label{PDP_individual_n}
 \widehat{\E}[\widetilde{\by}^{[j]}_n] = \left[\frac{\sum_{n = 1}^{N}{\Psi_W(\Tilde{\bx}_n^{[j]}(1))v_n}}{N}, \ldots, \frac{\sum_{n = 1}^{N}{\Psi_W(\Tilde{\bx}_n^{[j]}(K_j))v_n}}{N}  \right].
\end{equation} 
This can be interpreted as an empirical average, averaging over all instances $1\le n \le N$.

\section{ICEnet} \label{ICEnet_section}
\subsection{Description of the ICEnet}

\begin{figure}[htb!] 
	\begin{center}
		\begin{minipage}{1\textwidth}
			\begin{center}
				\includegraphics[width=\linewidth, height =1\linewidth]{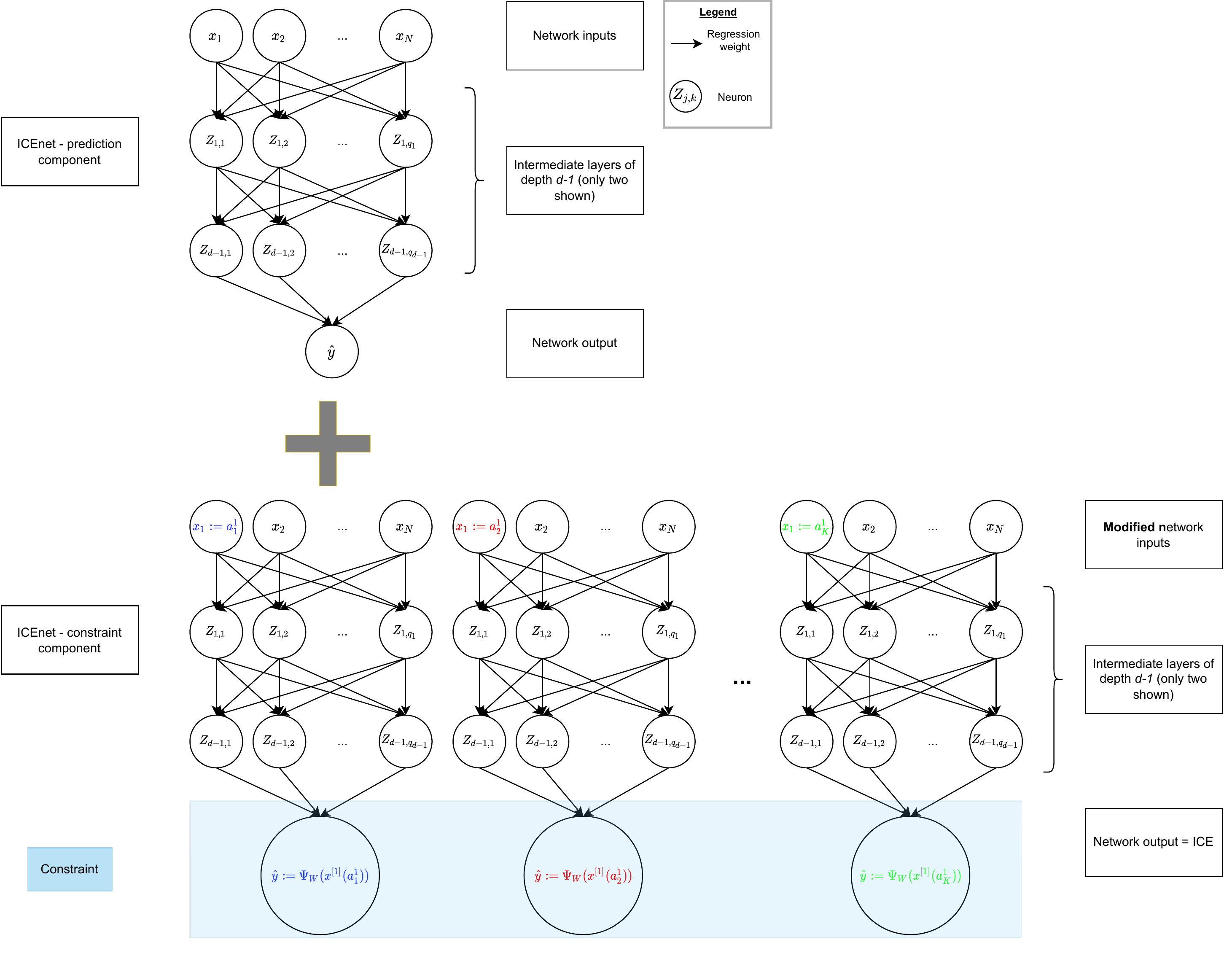}
			\end{center}
		\end{minipage}
	\end{center}
	\caption{Diagram explaining the ICEnet. The same neural network $\Psi_W$ is used to produce both the predictions from the model, as well as to create predictions based on pseudo-data. These latter predictions are constrained, ensuring that the outputs of the ICEnet vary smoothly or monotonically with changes in the input variables $\bx$. In this graph, we are varying variable $x_1$  to produce the ICEnet outputs which are $\Psi_W(\Tilde{\bx}^{[1]}(\cdot))$.}
	\label{icenet_diag}
\end{figure}

We start with a colloquial description of the ICEnet proposal before defining it rigorously. The main idea is to augment the observed data $(y_n, \bx_n, v_n)_{n=1}^N$ with pseudo-data, so that output equivalent to that used for the ICE interpretability technique \eqref{ICE_individual_n} is produced by the network, for each variable that requires a smoothness or monotonicity constraint. This is done be creating pseudo-data that varies for each possible value of the variables to be constrained. For continuous variables, we use quantiles of the observed values of the continuous variables to produce the ICE output while, for categorical variables, we use each level of the categories. The same neural network is then applied to the observed data, as well as each of the pseudo-data. Applying the network to the actual observed data produces a prediction that can be used for pricing, whereas applying the network to each of the pseudo-data produces outputs which vary with each of the covariates which need smoothing or require monotonicity to be enforced. The parameters of the network are trained (or fine-tuned) using a compound loss function: the first component of the loss function measures how well the network predicts the observations $y_n$, $1\le n \le N$. The other components ensure that the desired constraints are enforced on the ICEs for the constrained variables. After training, the progression of predictions of the neural network will be smooth or monotonically increasing with changes in the constrained variables. A diagram of the ICEnet is shown in Figure \ref{icenet_diag}.

\subsection{Definition of the ICEnet}

To define the ICEnet, we assume that some of the variables comprising the covariates $X\in \R^{N\times(q_r+q_c)}$ have been selected as requiring smoothness and monotonicity constraints. We collect all those variables requiring smoothness constrains into a set ${\cal S}\subset \{1,\ldots, q_r+q_c\}$ with $S$ members of the set and those variables requiring monotonicity constraints into another set ${\cal M}\subset \{1,\ldots, q_r+q_c\}$, with $M$ members of the set. As we have discussed above, we will rely on FCNs with appropriate pre-processing of the input data $\bx$ for predicting the response $y$ with $\widehat{y}$. 

For each instance $n \in \{1, \ldots, N\}$, the ICEnet is comprised of two main parts. The first of these is simply a prediction $\widehat{y}_n=\Psi_W(\bx_n)v_n$ of the outcome $y_n$ based on covariates $\bx_n$. For the second part of the ICEnet, we now will create the pseudo-data using the definitions from Section \ref{ICE_method} that will be used to create the ICE output of the network for each of the columns requiring constraints. Finally, we assume that the ICEnet will be trained with a compound loss function $L$, that balances the good predictive performance of the predictions $\widehat{y}_n$ together with satisfying constraints to enforce smoothness and monotonicity. 

The compound loss function $L$ of the ICEnet consists of three summands, i.e., $L = L_1 + L_2 + L_3$. The first of these is set equal to the deviance loss function $L^D$ that is relevant for the considered regression problem, i.e.,  $L_1 \seteq L^D(y_n, \widehat{y}_n)$. 

The second of these losses is a smoothing constraint applied to each covariate in the set ${\cal S}$. For smoothing, actuaries have often used the Whittaker--Henderson smoother \cite{Whittaker, Henderson}, which we implement here as the square of the third difference of the predictions $\widetilde{\by}^{[j]}_n$ given in \eqref{ICE_individual_n}. We define the difference operator of order 1 as
\begin{equation*}
\Delta^1(\Psi_W(\Tilde{\bx}_n^{[j]}(u))) = \Psi_W(\Tilde{\bx}_n^{[j]}(u)) - \Psi_W(\Tilde{\bx}_n^{[j]}(u-1)),
\end{equation*} 
and the difference operator of a higher order $\tau \geq 1$ recursively as 
\begin{equation*}
\Delta^\tau(\Psi_W(\Tilde{\bx}_n^{[j]}(u))) = \Delta^{\tau-1}(\Psi_W(\Tilde{\bx}_n^{[j]}(u))) - \Delta^{\tau-1}(\Psi_W(\Tilde{\bx}_n^{[j]}(u-1))).
\end{equation*} 
Thus, the smoothing loss $L_2$ for order 3 is defined as
\begin{equation}\label{loss smoothing definition}
L_2(n) \,\defeq\, \sum_{j \in {\cal S}} \sum_{u=4}^{K_j} \lambda_{s_j}\left[\Delta^3(\Psi_W(\Tilde{\bx}_n^{[j]}(u)))\right]^2,
\end{equation} 
where $\lambda_{s_j}\ge 0$ is the penalty parameter for the smoothing loss for the $j$-th member of the set ${\cal S}$.

Finally, to enforce monotonicity, we add the absolute value of the negative components of the first difference of  $\widetilde{\by}^{[j]}_n$ to the loss, i.e., we define $L_3$ as:
\begin{equation}\label{loss monotone definition}
L_3(n) \,\defeq \,\sum_{j \in {\cal M}} \sum_{u=2}^{K_j} \lambda_{m_j}\max\left[\delta_j\Delta^1(\Psi_W(\Tilde{\bx}_n^{[j]}(u))),0\right],
\end{equation} 
where $\lambda_{m_j}\ge 0$ is the penalty parameter for the smoothing loss for the $j$-th member of the set ${\cal M}$, and
where $\delta_j = \pm 1$ depending on whether we want to have a monotone increase ($-1$) or decrease ($+1$) in the $j$-the variable of ${\cal S}$.

\begin{ass}[ICEnet architecture] \label{ICEnet}
Assume we have independent responses and covariates $(y_n, \bx_n^{[r]}, \bx_n^{[c]}, v_n)_{n=1}^N$ as defined in Section \ref{supervised}, and we have selected some covariates requiring constraints into sets ${\cal S}$ and ${\cal M}$. Assume we have a neural network architecture $\Psi_W$ as defined in \eqref{FFN generic} having network weights $W$. For the prediction part of the ICEnet, we use the following mapping provided by the network
\begin{equation*}
    \bx_n ~\mapsto~ \widehat{y}_n = \Psi_W(\bx_n)v_n,
\end{equation*} 
which produces the predictions required for the regression task. In addition, the ICEnet also produces the following predictions made with the same network on pseudo-data defined by 
\begin{equation}\label{ICEnet_assumption}
  \Tilde{\bx}_n ~ \mapsto ~ \left[\widetilde{\by}_n^{[s_1]}, \ldots, \widetilde{\by}_n^{[s_S]}, 
  \widetilde{\by}_n^{[m_1]}, \ldots, \widetilde{\by}_n^{[m_M]} \right],
\end{equation} 
where all definitions are the same as those defined in Section \ref{ICE_method} for $s_l \in {\cal S}$ and $m_l \in {\cal M}$, respectively. Finally, to train the ICEnet, we assume that a compound loss function, applied to each observation $n \in \{1,\ldots, N\}$ individually is specified as follows
\begin{equation}\label{ICEnet_loss}
  L(n) \,\defeq \, L^D(y_n, \widehat{y}_n) + L_2(n) + L_3(n),
\end{equation} 
for smoothing loss $L_2(n)$ and montonicity loss $L_3(n)$
given in \eqref{loss smoothing definition} and \eqref{loss monotone definition}, respectively,
for non-negative penalty parameters collected into a vector $\lambda = (\lambda_{s_1}, \ldots,\lambda_{s_S}, \lambda_{m_1}, \ldots,\lambda_{s_M})^\top$.
\end{ass}

We briefly remark on the ICEnet architecture given in Assumptions \ref{ICEnet}.

\begin{rems}\normalfont
\begin{itemize}

\item To produce the ICE outputs from the network under Assumptions  \ref{ICEnet}, we apply the same network  $\Psi_W(\cdot)$ multiple times to pseudo-data that have been modified to vary for each value that a particular variable can take, see
\eqref{ICEnet_assumption}. Applying the same network multiple times is called a point-wise neural network in \citeA{Vaswani2017}, and it is called a \texttt{time-distributed} network in the \texttt{Keras} package; see \citeA{Keras}. This application of the same network multiple times is also called a one-dimensional convolutional neural network.

\item Common actuarial practice when fitting general insurance pricing models is to consider PDPs to understand the structure of the fitted coefficients of a GLM and smooth these manually. Since we cannot smooth the neural network parameters $W$ directly, we have rather enforced constraints \eqref{loss smoothing definition} for the ICE produced by the network $\Psi_W(\cdot)$; this in particular applies if we smooth ordered categorical variables.

\item Enforcing the same constraints as those in \eqref{ICEnet_loss} in a GLM will automatically smooth/constrain the GLM coefficients similar to LASSO and fused LASSO regularization; we refer to \citeA{lasso}. We also mention that similar ideas have been used in enforcing monotonicity in multiple quantile estimation; see \citeA{Kellner}.

\item Since the PDP of a model is nothing more than the average over the ICE outputs of the model for each instance $n$,  by enforcing the constraints for each instance also enforces the constraints on the PDP. Moreover, the constraints in \eqref{ICEnet_loss} could be applied in a slightly different manner, by first estimating a PDP using \eqref{PDP_individual_n} for all of the observations in a batch, then applying the constraints to the estimated PDP.

\item We have used a relatively simple neural network within the ICEnet; of course, more complex network architectures could be used.

\end{itemize}
\end{rems}

\section{Applying the ICEnet} \label{applying}

\subsection{Introduction and exploratory analysis}

In this section, we apply the ICEnet to the French Motor Third Party Liability (MTPL) dataset included in the \texttt{CASdatasets} package in the \texttt{R} language \cite{dutang2020package}, with the aim of predicting claims frequency. We follow the same data pre-processing, error correction steps and splitting of the data into learning and testing sets (in a 9:1 ratio), as those described in Appendix B of \citeA{Wuthrich2021}, to which we also refer for an exploratory data analysis of this dataset. Table \ref{mtpl_eda} shows the volumes of data in each of the learning and testing sets, illustrating that the observed claims rate is quite similar between the two sets.

\begin{longtable}{lrrrr}
\toprule
Set & $N$ & Exposure & Claims & Frequency \\ 
\midrule
learn & $610,206$ & $322,392$ & $23,737$ & $0.0736$ \\ 
test & $67,801$ & $35,967$ & $2,644$ & $0.0735$ \\ 
\bottomrule
\caption{Number of records, total exposure and observed claims frequency in the French MTPL dataset, split into learning and testing sets}
\label{mtpl_eda}
\end{longtable}

\begin{figure}[htb!] 
	\begin{center}
		\begin{minipage}{0.9\textwidth}
			\begin{center}
				\includegraphics[width=\linewidth]{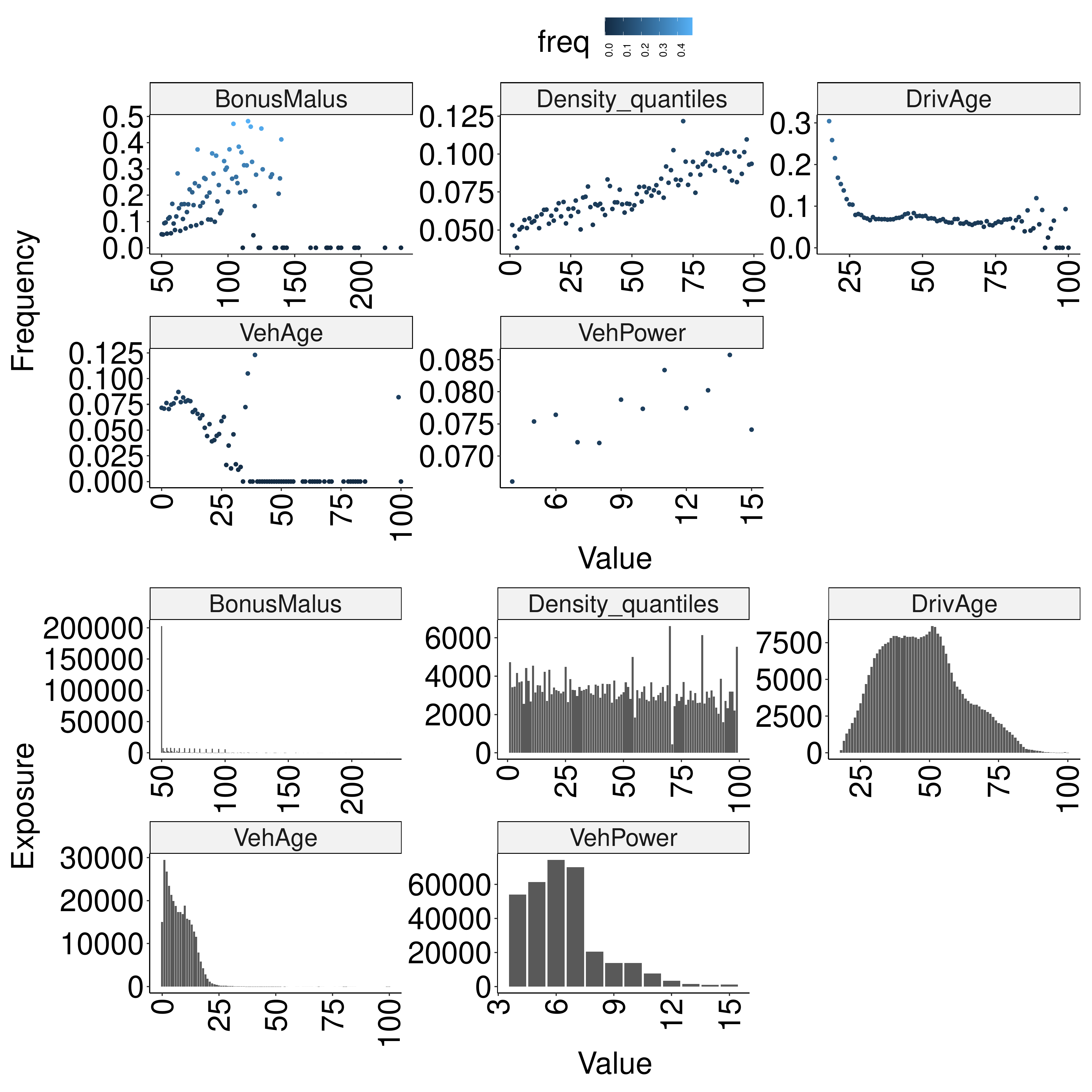}
			\end{center}
		\end{minipage}
	\end{center}
	\caption{Empirical claims frequency (top panel) and observed exposures (bottom panel) in the French MTPL dataset for each of the Bonus-Malus Level, Density, Driver Age, Vehicle Age and Vehicle Power covariates (univariate analysis only), learning set only. Note that the $y$-scales for each variable are not comparable.}
	\label{eda}
\end{figure}

The MTPL data contain both unordered and ordinal categorical covariates, as well as continuous covariates. To apply the ICEnet, we select all of the ordinal categorical and continuous covariates in the dataset for constraints; these are the Bonus-Malus Level, Density, Driver Age, Vehicle Age and Vehicle Power fields. Figure \ref{eda} shows the empirical claims frequency in the learning set, considering each fields in turn, i.e., this is a univariate (marginal) analysis. Note that for the continuous density variable, we are showing the frequency and exposure calculated at the percentiles of this variable. It can be seen that the empirical frequency varies with each variable somewhat erratically, especially in those parts of the domain of the variable when there is small exposure. In particular, for the bonus-malus level and density variables, there is quite a significant vertical spread of empirical frequencies, indicating that the univariate analysis presented here may not suitably capture the relationships between the covariates and the response, i.e., we need a more complex model to make predictions on this data.

In what follows, we impose the following constraints on these variables. Smoothing constraints are applied to all five of the variables. Among these five smoothed variables, ensuring that smoothness is maintained for driver age, vehicle age and bonus-malus level will be the most important constraints needed in this model from a commercial perspective, since the values of these variables will likely be updated each year that a policy is in force. Since we expect that claims frequency will increase with increasing bonus-malus scores, density and vehicle power, we also apply constraints to ensure that the model predicts monotonically increasing claims frequency for each of these variables. Table \ref{lambda_tab} shows these penalty parameters for each variable. These constraint values were selected heuristically by experimenting with different values until acceptably smooth and monotonic ICE outputs were produced, however, the impact on predictive performance was not considered when selecting these. Note that the table includes a direction column to show that we are enforcing monotonically increasing constraints to $\delta_j=-1$; setting the direction parameter to $\delta_j=1$ would produce monotonically decreasing constraints (see Listing \ref{CombinedLoss} in Appendix \ref{code}). Of course, these parameters could also be selected to produce optimal predictive performance using, for example, an out-of-sample validation set, or $K$-fold cross-validation.

\afterpage{
  \clearpage
\begin{longtable}{lrrr}
\toprule
Covariate & Smoothing Constraint & Monotonicity Constraint & Direction \\ 
\midrule
Driver Age & 10 & 0 & $-1$ \\ 
Vehicle Age & 1 & 0 & $-1$ \\ 
Bonus Malus & 1 & 100 & $-1$ \\ 
Density & 1 & 100 & $-1$ \\ 
Vehicle Power & 1 & 100 & $-1$ \\ 
\bottomrule
\caption{Smoothing and monotonicity constraints applied within the ICEnet}
\label{lambda_tab}
\end{longtable}
}

\subsection{Fitting the ICEnet} \label{fitting_ICEnet}

We select a $d=3$ layer neural network for the FCN component of the ICEnet, with the following layer dimensions $(q_1 = 32, q_2 = 16, q_3 = 8)$ and set the activation function $\phi(\cdot)$ to the Rectified Linear Unit (ReLU) function. The link function $g(\cdot)$ was chosen to be the exponential function (which is the canonical link of the Poisson model). To perform early-stopping to regularize the network, we  further split the learning set into a new learning set ${\cal L}$, containing 95\% of the original learning set, and a small validation set ${\cal V}$, containing 5\% of the original learning set. We assign an embedding layer to each of the unordered categorical variables (Vehicle Gas, Vehicle Brand, Region and Area Code), and, for simplicity, select the dimension of the real-valued embedding to be $b=5$ for each of these variables. For the rest of the variables, we apply mix-max scaling (see Section \ref{processing}) and then directly input these into the FCN; this explains how the vector of covariates $\bx$ has been constructed.

For comparison with the ICEnet, we begin with training several simple FCNs for comparison; 10 training runs of these were performed. This was fit using the Adam optimizer for 50 epochs, using a learning rate of $0.001$ and a batch size of $1024$ observations. The network training was early stopped by selecting the epoch with the best out-of-sample score on the validation set ${\cal V}$. To train this FCN network, we use only the deviance loss $L^D$, i.e., the first component of the loss function \eqref{ICEnet_loss}. Since neural network training involves some randomness arising from the random initialization of the network weights $W$ and the random selection of training data to comprise each mini-batch, the results of each training run will vary in terms of predictive performance; furthermore the exact relationships learned between the covariates and the predictions will vary by training run. In the first two lines of Table \ref{tab_res1}, we show the average of the Poisson deviance loss and the Poisson deviance loss produced using the nagging predictor of \citeA{richman2020nagging}, which is the predictor produced by averaging over the outputs of several neural network training runs. The standard deviation of the Poisson deviance loss across training runs is shown in the table in parentheses.

The ICEnet was fit using the \texttt{Keras} in the \texttt{R} programming language;\footnote{The FCN networks were fit on an AMD Ryzen 7 8 Core Processor with 32GB of RAM. The ICEnets were fit on the same system, using an Nvidia RTX 2070 GPU and the GPU-enabled version of Tensorflow.} for the {\tt Keras} package see \citeA{Keras}. The implementation of the ICEnet is explained in more detail in Appendix \ref{code} in the supplementary material. Similar to the FCNs, 10 training runs of the ICEnet were performed. Since the ICEnet requires many outputs to be produced for each input to the network, the computational burden of fitting this model is quite heavy, nonetheless, this can be done easily using a Graphics Processing Unit (GPU) and the relevant GPU optimized versions of deep learning software. The ICEnet results are on Lines 3-4 of Table \ref{tab_res1}, showing the average results and the nagging predictor, respectively. We note that training runs of the FCN and ICEnet take about 3 and 12 minutes to complete, respectively, the former running without a GPU and the latter running on a GPU.

\afterpage{
  \clearpage
\begin{longtable}{l|rr|rr}
\toprule
Description & Learn && Test & \\ 
\hline
FCN & 0.2381& (0.000211) & 0.2387& (0.000351) \\ 
FCN (nagging) & 0.2376 && 0.2383& \\ \hline
ICEnet & 0.2386& (0.000180) & 0.2388& (0.000236) \\ 
ICEnet (nagging) & 0.2384 && 0.2385& \\ 
\bottomrule
\caption{Poisson deviance loss for the FCN and the ICEnet, learning and testing losses. For multiple runs, the average and standard deviation (in parentheses) are reported.}
\label{tab_res1}
\end{longtable}
}

It can be observed that adding smoothing and monotonicity constraints to the ICEnet has resulted in only marginally worse performance of the model on both the training and testing set, compared to the FCN results. Interestingly, whereas the FCN has a relatively large performance gap between the training and testing sets (both on average and when comparing the nagging predictor), the ICEnet has a much smaller gap, implying that these constraints are regularizing the network, at least to some extent. Finally, while the nagging predictor improves the performance of both the FCN and the ICEnet, the performance improvement is somewhat larger for the former than the latter, which can be explained by the nagging predictor performing regularization over the different fitting runs.

We approximately decompose the reduction in performance between the smoothness constraints on the one hand, and the monotonicity constraints on the other, by refitting the ICEnet with the smoothing constraints only, and then with the monotonicity constraints only. These results are shown in Table \ref{tab_res2}.

\begin{longtable}{ll|rr|rr}
\toprule
Description & Constraints & Learn && Test& \\ 
\midrule\hline
ICEnet & Smoothing + Monotonicity & 0.2386 &(0.000180) & 0.2388 &(0.000236) \\ 
ICEnet & Smoothing & 0.2385 &(0.000423) & 0.2388& (0.000385) \\ 
ICEnet & Montonicity & 0.2384 &(0.000214) & 0.2385& (0.000140) \\ \hline
ICEnet (nagging) & Smoothing + Monotonicity & 0.2384 && 0.2385& \\ 
ICEnet (nagging) & Smoothing & 0.2382 && 0.2385& \\ 
ICEnet (nagging) & Monotonicity & 0.2381 && 0.2382& \\ 
\bottomrule
\caption{Poisson deviance loss for the ICEnet, learning and testing sets. For multiple runs, the average and standard deviation (in parentheses) are reported. Constraints are varied as described in the relevant column.}
\label{tab_res2}
\end{longtable}

The results show that the smoothing constraints are the primary cause of the decline in performance of the ICEnet, compared with the FCN. The monotonicity constraints, on the other hand, improve the out-of-sample performance of the ICEnet to surpass the FCN, whether on average, or when using the nagging predictor. Furthermore, the small gap between the training and testing results for the ICEnet with monotonicity constraints only, shows that these successfully regularize the model. These results suggest that adding monotonicity constraints to actuarial deep learning models based on expert knowledge can lead to enhanced performance. Whether the somewhat decreased performance of a model that produces smoothed outputs is acceptable, will depend on whether these constraints are necessary for commercial purposes. Moreover, the values of the constraint parameters could be (further) varied so that an acceptable trade-off of smoothness and performance is achieved; for an example of this, see Section \ref{varying}, below.

\subsection{Exploring the ICEnet predictions}
\begin{figure}[htb!] 
	\begin{center}
		\begin{minipage}{0.9\textwidth}
			\begin{center}
				\includegraphics[width=\linewidth]{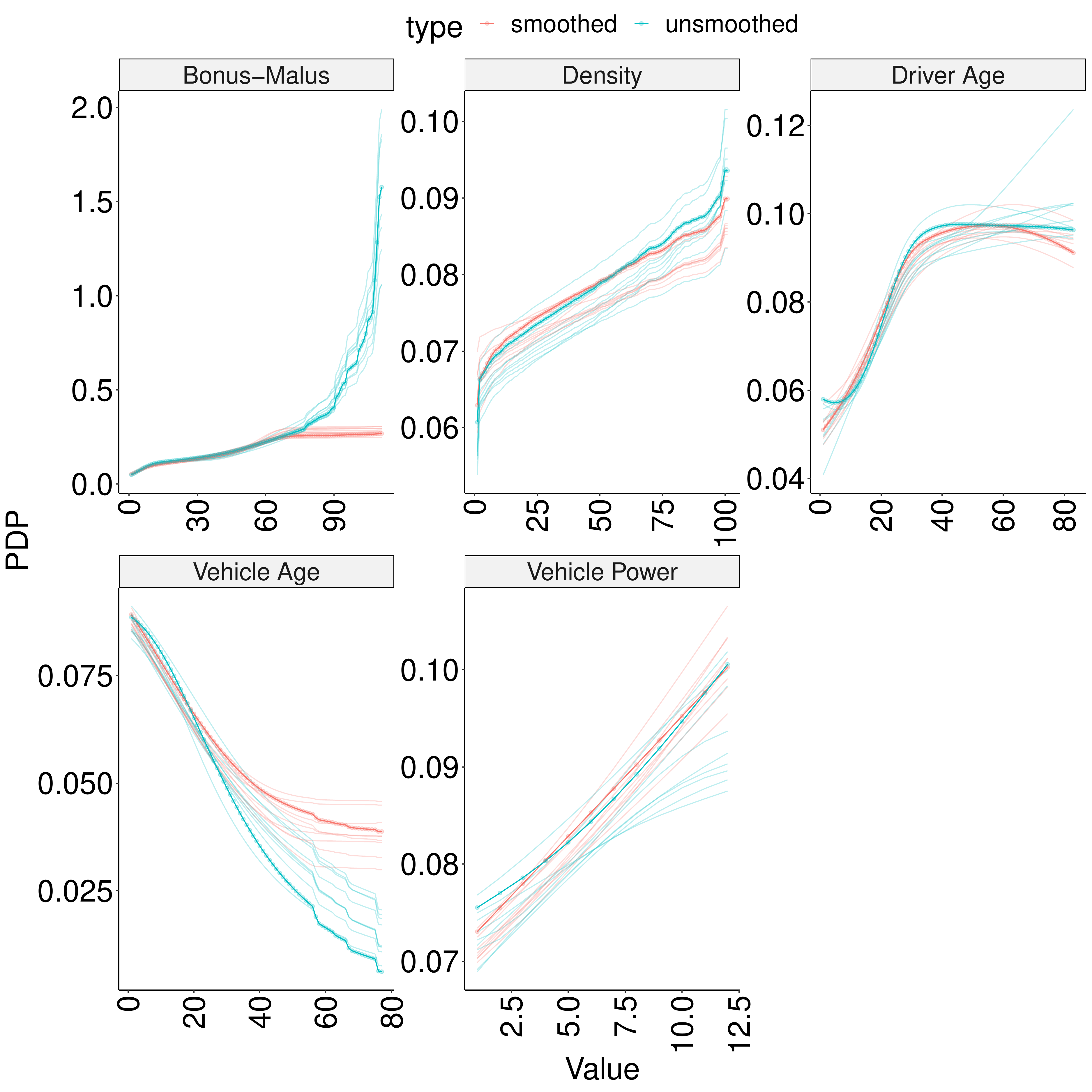}
			\end{center}
		\end{minipage}
	\end{center}
	\caption{PDPs for each of the Bonus-Malus Level, Density, Driver Age, Vehicle Age and Vehicle Power fields shown in separate panels, test set only. Blue lines are PDPs from the FCNs (unsmoothed) and red lines are PDPs from the ICEnet (smoothed). Bold lines relate to the PDPs from the first of 10 runs; the lighter lines relate to the remaining runs. Note that the scale of the $y$-axis varies between each panel.}
	\label{pdp}
\end{figure}

We start with considering the global impact of applying constraints within the ICEnet by considering the PDPs derived from the FCN and ICEnet models in each of the 10 training runs of these models. The ICE outputs were constrained when fitting the network by using loss function \eqref{ICEnet_loss} applied to each of the observations in the training set. The PDPs are shown in Figure \ref{pdp} with blue lines for FCN and red lines for ICEnet.  

The PDPs from the unconstrained FCNs models exhibit several undesirable characteristics: for the bonus-malus level, density and vehicle age covariates, these exhibit significant roughness which would translate into unacceptable changes in rates charged to policyholders as, for example, bonus-malus scores are updated, or as the policyholder's vehicle ages. Also, in some parts of the PDPs, it can be observed that monotonicity has not been maintained by the FCNs. Finally, the PDPs for the driver age and vehicle power covariates exhibit different shapes over the different runs, meaning that the FCNs have learned different relationships between the covariates and the predictions. On the other hand, the PDPs from the ICEnets are significantly smoother for all of the variables and runs, appear to be monotonically increasing in all cases and finally, exhibit more similar shapes across the different runs. Interestingly, the relationships learned by the constrained ICEnets are quite different for the bonus-malus level, vehicle age and driver age variables than those learned by the FCN, with the PDPs of the ICEnet for the first two variables arguably more commercially reasonable than the those from the FCN.

\begin{figure}[htb!] 
	\begin{center}
		\begin{minipage}{0.9\textwidth}
			\begin{center}
				\includegraphics[width=\linewidth]{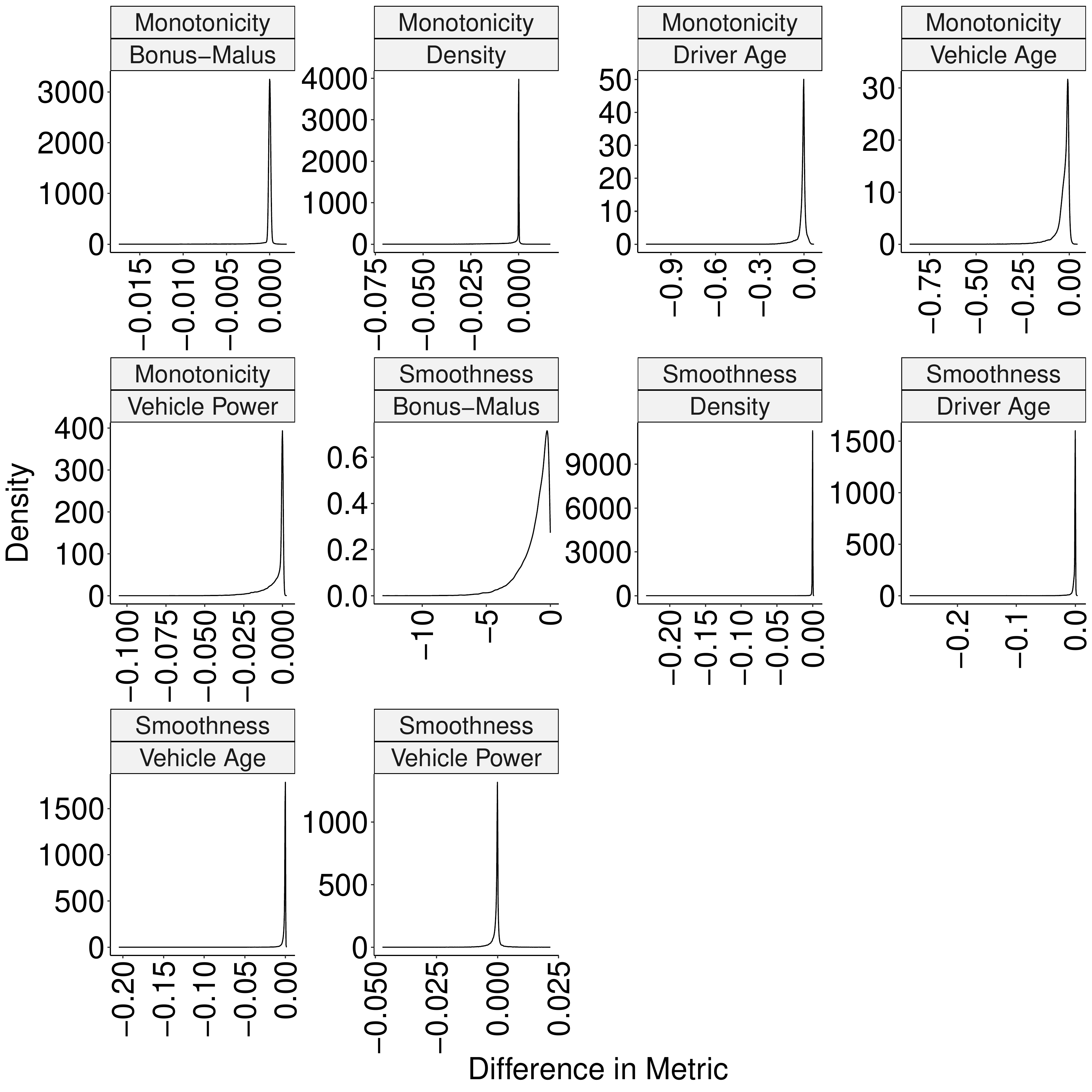}
			\end{center}
		\end{minipage}
	\end{center}
	\caption{Density plots of the difference between the monotonicity and smoothness components of the ICEnet loss function \eqref{ICEnet_loss} evaluated for each observation in the test set.}
	\label{ice_dens}
\end{figure}

Focusing on the first training run of the FCNs and the ICEnets, we now show some examples of the effect of the constraints on individual observations. To estimate the monotonicity and smoothness of the ICE outputs of the FCN and ICEnets, these components of the ICEnet loss function \eqref{ICEnet_loss} were evaluated for each observation in the test set. Figure \ref{ice_dens} shows density plots of the difference (smoothed model minus unsmoothed model) between these scores for each of these models, for each variable separately.

All of these densities have a long left tail, meaning to say, that the ICE outputs produced by the ICEnet are significantly more monotonic and smooth than those produced by the FCN. Also, the densities usually peak around zero, meaning to say that adding the constraints has generally not significantly ``damaged" outputs from the FCN that already satisfied the monotonicity or smoothness constraints. For the densities of the monotonicity scores for the bonus-malus level and density variables, it can be seen that there is a short right tail as well, which we have observed to occur when the original FCN model produces some outputs that are already monotonic and these are altered somewhat by the ICEnet.

\begin{figure}[htb!] 
	\begin{center}
		\begin{minipage}{0.9\textwidth}
			\begin{center}
				\includegraphics[width=\linewidth]{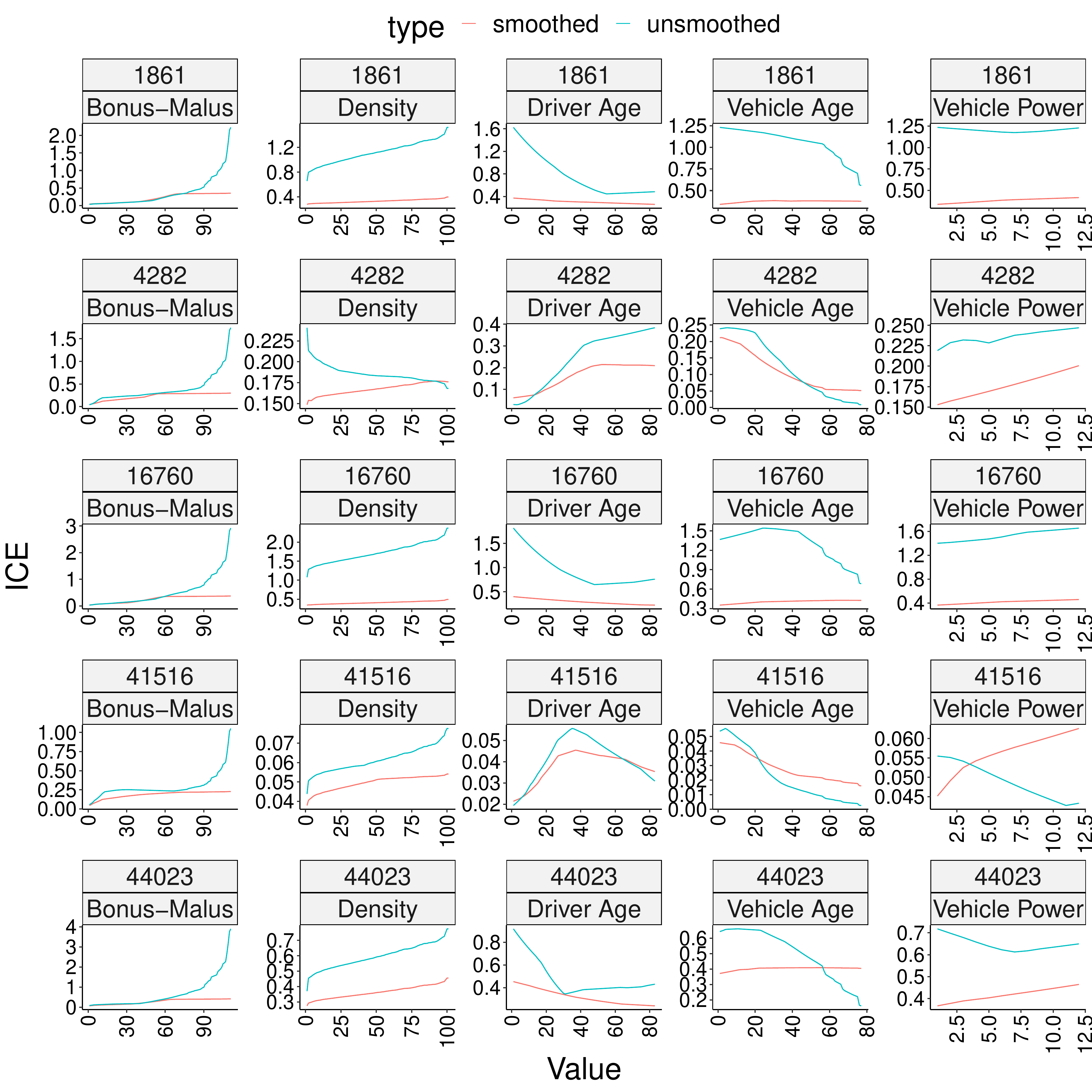}
			\end{center}
		\end{minipage}
	\end{center}
	\caption{ICE plots of the output of the FCN and the ICEnet for instances $n$ chosen to be the least monotonic based on the monotonicity score evaluated for each instance in the test set on the outputs of the FCN. Note that the smoothed model is the ICEnet and unsmoothed model is the FCN.}
	\label{ices_mon}
\end{figure}

\begin{figure}[htb!] 
	\begin{center}
		\begin{minipage}{0.9\textwidth}
			\begin{center}
				\includegraphics[width=\linewidth]{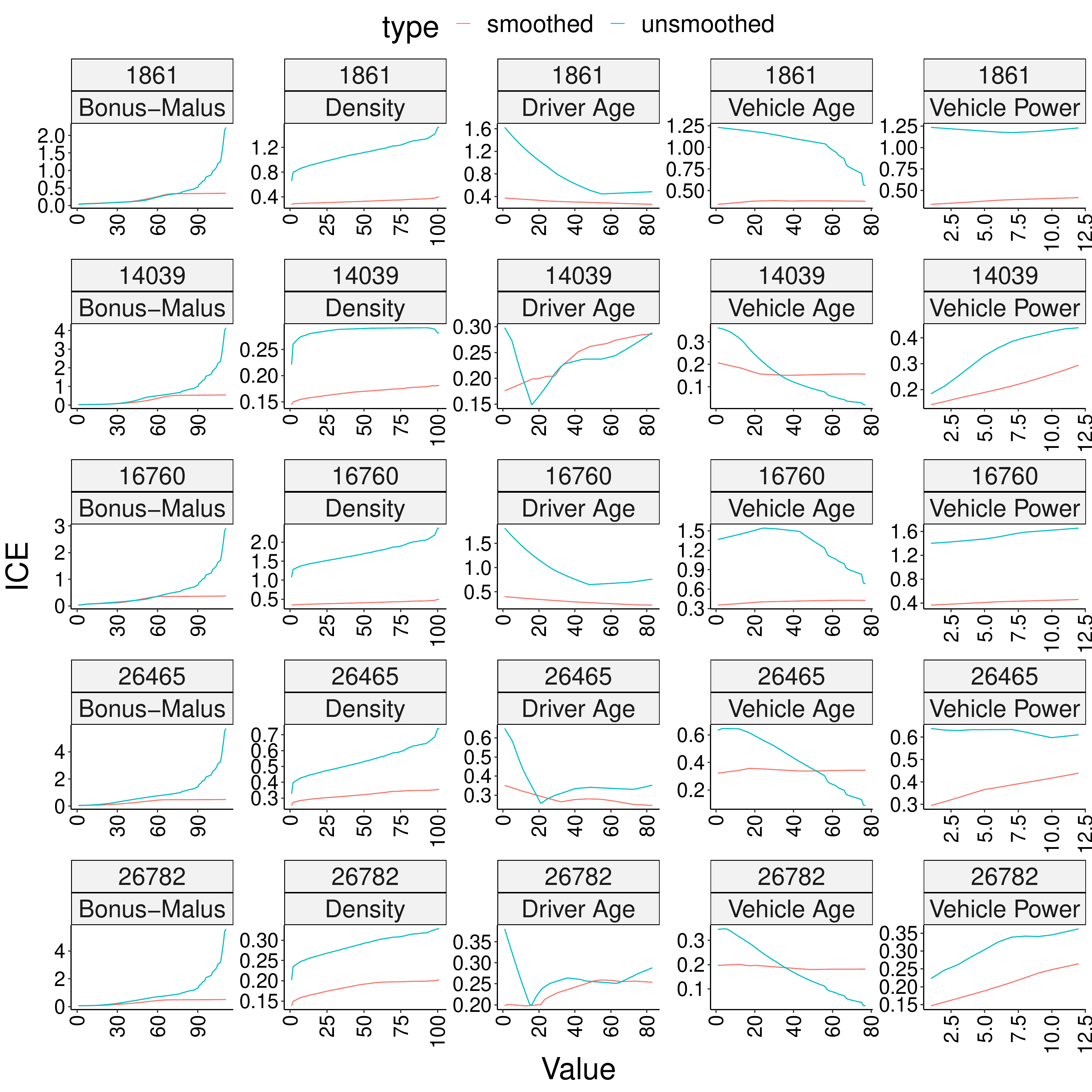}
			\end{center}
		\end{minipage}
	\end{center}
	\caption{ICE plots of the output of the FCN and the ICEnet for instances $n$ chosen to be the least smooth based on the smoothness score evaluated for each instance in the test set on the outputs of the FCN. Note that the smoothed model is the ICEnet and unsmoothed model is the FCN.}
	\label{ices_smooth}
\end{figure}

Figures \ref{ices_mon} and  \ref{ices_smooth} show ICE plots for several instances $n$, which have been specially chosen as those that are the least monotonic and smooth ones, based on the monotonicity and smoothness scores evaluated for each observation in the test set on the outputs of the FCN only. These figures show that, in general, the ICE plots of predictions from the ICEnet are more reasonable. For example, for instance $n=4282$ in Figure \ref{ices_mon}, it can be seen that the FCN produces predictions that decrease with density, whereas the opposite trend is produced by the ICEnet; another example is instance $n=41516$ where the FCN predictions decrease with vehicle power and the ICEnet reverses this. A nice side effect of the constraints is that some of the exceptionally large predictions produced by the FCN, for example, for instance $n=16760$ in Figure \ref{ices_smooth}, are reduced significantly by the ICEnet; in general it would be highly unlikely to underwrite policies with such an elevated frequency of claim.

\subsection{Varying the ICEnet constraints} \label{varying}

\begin{figure}[htb!] 
	\begin{center}
		\begin{minipage}{0.9\textwidth}
			\begin{center}
				\includegraphics[width=\linewidth]{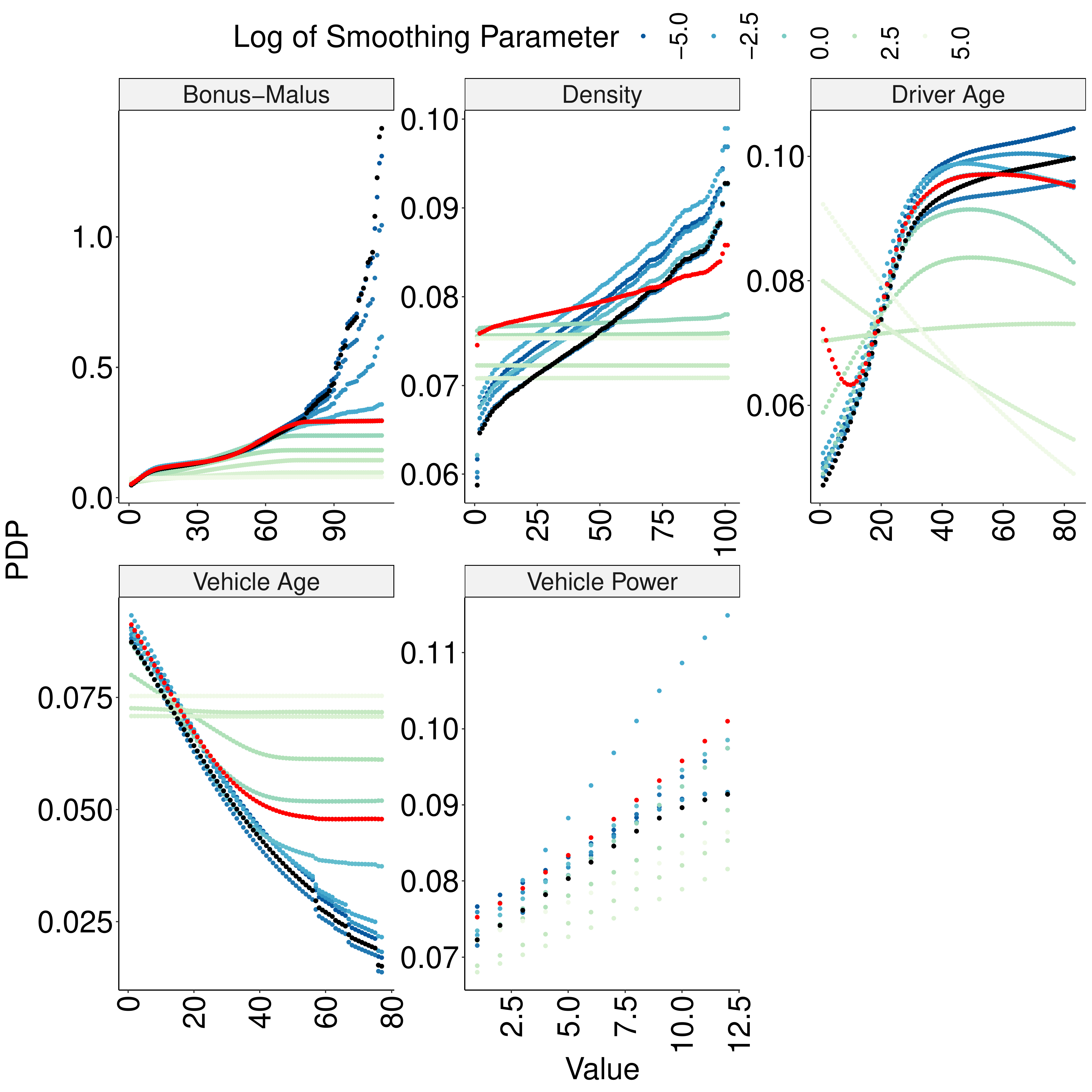}
			\end{center}
		\end{minipage}
	\end{center}
	\caption{PDPs of the meta-model and ICEnets with the constraint parameters varied from $10^{-5}$ to $10^5$. The PDPs of the meta-model are in black and the PDP of the optimal model - with smoothing parameters set equal to $1$ -  based on the validation set is shown in red.}
	\label{pdps_smoothing}
\end{figure}

We end this section by investigating how the PDPs of the ICEnets change as the strength of the constraint parameters are varied over a wide range. To provide a baseline starting point, a meta-network (or, using the terminology of machine learning, a distillation network, \citeA{hinton2015distilling}) was fit to the nagging predictor derived from the FCNs in Section \ref{fitting_ICEnet}; this was done by fitting exactly the same FCN as before, with the same covariates, but with the aim of approximating the nagging predictor derived using the 10 FCN predictions. Then, the weights of the meta-network were used as a starting point for fitting an ICEnet, where the strength of the constraints was varied over a wide range from $10^{-5}$ to $10^5$. The results of fitting these models, as well as the PDPs from the meta-model and the ICEnets are shown in Table \ref{tab_smoothing} and Figure \ref{pdps_smoothing}, respectively. The minimum value of the test set Poisson deviance loss is reached when setting the value of the constraints equal to $10^{-3}$, however, the validation set minimum is reached when setting the value of the constraints to unity; in this case the validation set identifies an almost optimal value of the constraints as measured by the test set performance. It can also be seen that the highest value of the constraints leads to models with a poor test and validation set performance. The PDPs show that the ICEnet with the optimal value of the constraints maintains many of the features of the PDPs of the meta-model, while being smoother and less extreme than the meta-model. The ICEnets with the most extreme values of the constraint parameters become much flatter for the bonus-malus level, vehicle age and density variables, whereas the PDPs for driver age and vehicle power maintain significant variation over the range of these variables.

\afterpage{
  \clearpage
\begin{longtable}{lrrr}
\toprule
Regularization Parameter ($\log_{10}$) & Learn & Validation & Test \\ 
\midrule
meta-model & $0.23835$ & $0.23232$ & $0.23829$ \\ 
-5 & $0.23838$ & $0.23276$ & $0.23856$ \\ 
-4 & $0.23833$ & $0.23245$ & $0.23847$ \\ 
-3 & $0.23791$ & $0.23243$ & $0.23837$ \\ 
-2 & $0.23831$ & $0.23263$ & $0.23855$ \\ 
-1 & $0.23816$ & $0.23241$ & $0.23856$ \\ 
\textbf{0} & $\textbf{0.23785}$ & $\textbf{0.23236}$ & $\textbf{0.23840}$\\ 
1 & $0.23924$ & $0.23332$ & $0.23947$ \\ 
2 & $0.24093$ & $0.23484$ & $0.24099$ \\ 
3 & $0.24341$ & $0.23684$ & $0.24375$ \\ 
4 & $0.24659$ & $0.23965$ & $0.24744$ \\ 
5 & $0.24903$ & $0.24212$ & $0.25018$ \\ 
\bottomrule
\caption{Poisson deviance loss for the ICEnet, learning, validation and testing sets. The magnitude of the constraints is varied as described in the first column; ``meta-model" denotes the results of the meta-model with no constraints applied.}
\label{tab_smoothing}
\end{longtable}
}

Here, we have considered quite simple variations of the constraint parameters by varying the constraints for all variables together; using other techniques, one could also attempt to find an optimal set of constraints where the values of these differed by variables and type of constraint.

\section{Local approximations to the ICEnet} \label{Local}
In the previous section, the constraints on the network have been enforced by creating ICE outputs from the FCN for each value that the covariates with constraints can take, see Assumptions \ref{ICEnet}. However, this creates quite some computational overhead when fitting the ICEnet. Thus, we now explore a local approximation to the ICEnet, where the ICE outputs from the FCN are created only for a small subset of the values that the covariates with constraints can take. In particular, we define a window size parameter $\omega$ and only produce the ICE outputs in a window of $\frac{\omega-1}{2}$ around the actual value of each covariate. To ensure that the third difference used for smoothing remains defined, we set the window parameter to have a value of $\omega = 5$ in what follows. To define the Local ICEnet approximation, all that is needed is to modify the ICEnet to produce outputs for an observation to return outputs only in the window around the actual observed value. Furthermore, for instances that are at the extremes of the observed value of $j$-th covariate, e.g., for the youngest drivers, we produce outputs from the ICEnet for only the first or last $\omega$ observations of the set $\{a^j_1,\ldots, a^j_{K_j} \}$. Since the Local ICEnet needs to produce significantly fewer outputs than the global ICEnet presented earlier, the computational burden of fitting this model is dramatically decreased and the model can be fit without a GPU.

A Local ICEnet was fit to the same data as above. To enforce the constraints more strongly over the range of the constrained variables, slightly stronger values for these were found to work well; these are shown in Table \ref{lambda_tab_local}. A Local ICEnet takes about 12 minutes to run without a GPU; this is similar to the global ICEnet running on a GPU.

\afterpage{
  \clearpage
\begin{longtable}{lrrr}
\toprule
Covariate & Smoothing Constraint & Monotonicity Constraint & Direction \\ 
\midrule
Driver Age & 200 & 0 & $-1$ \\ 
Vehicle Age & 10 & 0 & $-1$ \\ 
Bonus Malus & 10 & 200 & $-1$ \\ 
Density & 10 & 200 & $-1$ \\ 
Vehicle Power & 10 & 200 & $-1$ \\ 
\bottomrule
\caption{Smoothing and monotonicity constraints applied within the Local ICEnet}
\label{lambda_tab_local}
\end{longtable}
}

The results of fitting the Local ICEnet are shown in Table \ref{tab_res3}. In this case, the ICEnet outperforms the FCN, both on average and for the nagging predictor and, comparing these results to those in Table \ref{tab_res1}, it appears that the Local ICEnet allows constraints to be enforced with a smaller loss of predictive performance than the global ICEnet shown above.

\begin{longtable}{l|rr|rr}
\toprule
Description & Learn && Test& \\ 
\hline
FCN & 0.2381 &(0.000211) & 0.2387& (0.000351) \\ 
FCN (nagging) & 0.2376 && 0.2383& \\ \hline
ICEnet & 0.2385& (0.000301) & 0.2386 &(0.000215) \\ 
ICEnet (nagging) & 0.2381& & 0.2383 &\\ 
\bottomrule
\caption{Poisson deviance loss for an FCN and the Local ICEnet, learning and testing sets. For multiple runs, the average and standard deviation (in parentheses) are reported.}
\label{tab_res3}
\end{longtable}

We show plots of the outputs of the Local ICEnet in Appendix \ref{local_plots} of the supplementary material. From Figure \ref{pdp_window} it can be seen that the global effect of applying the Local ICEnet is quite similar to that of the global ICEnet, however, the PDPs shown in this figure are a little less smooth than those of the global ICEnet. In Figure \ref{ice_dens_window} it can be seen that the Local ICEnet approximation has been moderately successful at enforcing the required constraints with most of the variables showing improved monotonicity and smoothness scores, except for the bonus-malus level and density covariates. The examples shown in Figures \ref{ices_mon_window} and \ref{ices_smooth_window} illustrate that for the most extreme examples of the non-monotonic and rough results produced by the FCN, the Local ICEnet appears to produce outputs that are similar to the global ICEnet.

\section{Conclusions} \label{conclusions}
We have presented a novel approach, the ICEnet, to constrain neural network predictions to be monotonic and smooth using the ICE model interpretability technique. We have shown how the global ICEnet can be approximated using a less computationally intensive version, the Local ICEnet, that enforces constraints on locally perturbed covariates. Fitting these models to a real-world open-source dataset has shown that the monotonicity constraints enforced when fitting the ICEnet can improve the out-of-sample performance of actuarial models, while fitting models with a combination of smoothing and monotonic constraints allows the models to produce predicted frequencies of claim that accord with intuition and are commercially reasonable. In this work, we have focused on smoothing the predictions of networks with respect to individual observations; as we have noted above it is easy to rather impose constraints on the global behaviour of networks instead. Moreover, other formulations of smoothing constraints could be imposed, for example, instead of squaring the third differences of model outputs, which will lead to minimizing the larger deviations from smoothness, rather absolute differences could be smoothed. 

{\small
	
	\bibliographystyle{apacite}	
\bibliography{ref}	 

}

\clearpage
\pagebreak

 \appendix

 \begin{center}
   {\LARGE   {\sc Supplementary}}
   \end{center}

\section{Appendix: code} \label{code}

The ICEnet was implemented using the \texttt{R} language version of the \texttt{Keras} package \cite{Keras}.

\medskip 

The constraints used for the ICEnet are implemented using a custom layer, which is shown in Listing \ref{CombinedLoss}. This custom layer works by either estimating the squared third differences (smoothing) or the squared value of negative first differences (monotonicity) (or both of these) and then adds the value of these penalties to the loss of the model. The monotonicity constraint can also be reversed to enforce monotonically decreasing constraints.

\begin{lstlisting}[float=h!,frame=tb,caption={Code for the Combined Loss Layer in \texttt{R}.},label=CombinedLoss]
# Custom combined loss layer
CombinedLossLayer <- Layer(
  classname = "CombinedLossLayer",
  initialize = function(lambda_smooth = 0, lambda_mono = 0, direction = 1) {
    super()$`__init__`()
    self$lambda_smooth <- lambda_smooth
    self$lambda_mono <- lambda_mono
    self$direction <- direction
  },
  call = function(inputs, ...) {
    input_shape <- k_shape(inputs)
    #total_loss <- k_constant(0, dtype = 'float32')
    
    # Smoothing loss
    if (self$lambda_smooth > 0) {
      input1 <- tf$slice(inputs, begin = c(0L, 0L), size = c(-1L, input_shape[2] - 3L))
      input2 <- tf$slice(inputs, begin = c(0L, 1L), size = c(-1L, input_shape[2] - 3L))
      input3 <- tf$slice(inputs, begin = c(0L, 2L), size = c(-1L, input_shape[2] - 3L))
      input4 <- tf$slice(inputs, begin = c(0L, 3L), size = c(-1L, input_shape[2] - 3L))
      
      diff1 <- input2 - input1
      diff2 <- input3 - input2
      diff3 <- input4 - input3
      
      third_diff <- diff3 - 2 * diff2 + diff1
      
      squared_diff3 <- k_sum(k_square(third_diff), axis=2)
      smooth_loss <- k_mean(squared_diff3, axis=1) * self$lambda_smooth
      #total_loss = total_loss + smooth_loss
      self$add_loss(smooth_loss)
    }
    
    # Monotonicity loss
    if (self$lambda_mono > 0) {
      input1 <- tf$slice(inputs, begin = c(0L, 0L), size = c(-1L, input_shape[2] - 1L))
      input2 <- tf$slice(inputs, begin = c(0L, 1L), size = c(-1L, input_shape[2] - 1L))
      
      first_diff <- input2 - input1
      adjusted_diff <- first_diff * self$direction
      penalty <- k_sum(k_square(k_maximum(adjusted_diff, 0)), axis=2)
      mono_loss <- k_mean(penalty, axis=1) * self$lambda_mono
      #total_loss = total_loss + mono_loss
      self$add_loss(mono_loss)
    }    
    
    # Return the input tensor
    inputs
  }
)
\end{lstlisting}

\medskip

The input processing for the ICEnet is shown in the following two listings in this section. The first of these, Listing \ref{ICEnetpart1}, shows the input processing for the prediction part of the network and the second of these, Listing \ref{ICEnetpart2}, shows the input processing to produce the ICE output of the ICEnet. The main idea of Listing \ref{ICEnetpart2} is to define a matrix of covariates, where all of the covariates entered into the prediction network are held constant, and only those covariates which relate to the ICE outputs vary. These are varied in turn for each covariate for which an ICE must be produced.

\medskip

Finally, Listing \ref{ICEnetpart3} shows the rest of the ICEnet. A single FCN is used to derive predictions from the inputs. This same network is used to make predictions for the ICE outputs, using the \texttt{time\_distributed} layer. The constraints are then enforced on each of the ICE outputs. 

\begin{lstlisting}[float=h!,frame=tb,caption={Code for the Input Processing in \texttt{R}.},label=ICEnetpart1]
embed_layer <- function(input, input_dim, output_dim, name) {
  input %>% 
    layer_embedding(input_dim=input_dim, output_dim=output_dim, input_length=1, name=name) %>% 
    layer_flatten()
}
  
  # Define inputs
  inputs <- list(
    Exposure = layer_input(shape = 1, dtype = 'float32', name = 'Exposure'),
    VehGas = layer_input(shape = 1, dtype = 'int32', name = 'VehGas'),
    VehBrand = layer_input(shape = 1, dtype = 'int32', name = 'VehBrand'),
    Region = layer_input(shape = 1, dtype = 'int32', name = 'Region'),
    Area = layer_input(shape = 1, dtype = 'int32', name = 'Area'),
    DrivAge = layer_input(shape = 1, dtype = 'float32', name = 'DrivAge'),
    VehAge = layer_input(shape = 1, dtype = 'float32', name = 'VehAge'),
    BonMal = layer_input(shape = 1, dtype = 'float32', name = 'BonMal'),
    Density = layer_input(shape = 1, dtype = 'float32', name = 'Density'),
    VehPower = layer_input(shape = 1, dtype = 'float32', name = 'VehPower')
  )
  
  # Define embeddings
  embeddings <- list(
    VehGas_embed = embed_layer(inputs$VehGas, 2, 5, 'VehGas_embed'),
    VehBrand_embed = embed_layer(inputs$VehBrand, 11, 5, 'VehBrand_embed'),
    Region_embed = embed_layer(inputs$Region, 22, 5, 'Region_embed'),
    Area_embed = embed_layer(inputs$Area, 6, 5, 'Area_embed')
  )
  
  # Define reshape functions for ICEs
  embed_ice <- function(input, len) {
    if(len > 1){
      input %>% layer_reshape(target_shape = c(len, 1))
    }
    else {
      input %>% layer_reshape(target_shape = c(1))
    }
    
  }
  
  num_embeds <- list(
    DrivAge_embed = embed_ice(inputs$DrivAge, 1),
    VehAge_embed = embed_ice(inputs$VehAge, 1),
    BonMal_embed = embed_ice(inputs$BonMal, 1),
    Density_embed = embed_ice(inputs$Density, 1),
    VehPower_embed = embed_ice(inputs$VehPower, 1)
  )
\end{lstlisting}

\begin{lstlisting}[float=h!,frame=tb,caption={Code for the ICE Processing in \texttt{R}.},label=ICEnetpart2]
 # Define ICE embeddings
  ices <- list(
    DrivAge_ice = layer_input(shape = c(DrivAge_ice_len), dtype = 'float32', 
    name = 'DrivAge_ice'),
    VehAge_ice = layer_input(shape = c(VehAge_ice_len), dtype = 'float32', 
    name = 'VehAge_ice'),
    BonMal_ice = layer_input(shape = c(BonusMalus_ice_len), dtype = 'float32', 
    name = 'BonMal_ice'),
    Density_ice = layer_input(shape = c(Density_ice_len), dtype = 'float32', 
    name = 'Density_ice'),
    VehPower_ice = layer_input(shape = c(VehPower_ice_len), dtype = 'float32', 
    name = 'VehPower_ice')
  )
  
  ices_embed <- list(
    DrivAge_ice_embed = embed_ice(ices$DrivAge_ice, DrivAge_ice_len),
    VehAge_ice_embed = embed_ice(ices$VehAge_ice, VehAge_ice_len),
    BonMal_ice_embed = embed_ice(ices$BonMal_ice, BonusMalus_ice_len),
    Density_ice_embed = embed_ice(ices$Density_ice, Density_ice_len),
    VehPower_ice_embed = embed_ice(ices$VehPower_ice, VehPower_ice_len))
  
  embeds <- embeddings %>% unname %>% layer_concatenate()
  num_embeds_flat <- num_embeds %>% unname %>% layer_concatenate()
  main = list(embeds, num_embeds_flat) %>% layer_concatenate() %>%
  layer_reshape(target_shape = c(1,25)) 
  
  #### Define ICE models
  concat_embeds <- function(embeddings, embeds, ice_embeds, ice_len, ice_idx) {
    repeated_embeddings <- lapply(embeddings, function(x) x %>% layer_repeat_vector(ice_len))
    repeated_embeds <- lapply(embeds, function(x) x %>% layer_repeat_vector(ice_len))
    repeated_embeds[[ice_idx]] <- ice_embeds
    temp1 = repeated_embeddings %>% unname %>% layer_concatenate()
    temp2 = repeated_embeds %>% unname %>% layer_concatenate()
    list(temp1, temp2) %>% layer_concatenate()
  }
  
  # ICE names, lengths and indices
  ice_info <- list(
    DrivAge_ice_embed = list(len = DrivAge_ice_len, idx = 1),
    VehAge_ice_embed = list(len = VehAge_ice_len, idx = 2),
    BonMal_ice_embed = list(len = BonusMalus_ice_len, idx = 3),
    Density_ice_embed = list(len = Density_ice_len, idx = 4),
    VehPower_ice_embed = list(len = VehPower_ice_len, idx = 5)
  )
  
  # Concatenate ICE embeddings
  ice_embed_vectors <- list()
  for (name in names(ices_embed)) {
    ice_embed_vectors[[paste0(name)]] <-
      concat_embeds(embeddings, num_embeds, ices_embed[[name]],
      ice_info[[name]]$len, ice_info[[name]]$idx)
    }
\end{lstlisting}

\begin{lstlisting}[float=h!,frame=tb,caption={Code for the rest of the ICEnet model in \texttt{R}.},label=ICEnetpart3]
 #### define main model
  middle_in = layer_input(shape = 25, dtype = "float32")
  middle_layers = middle_in %>% 
    layer_dense(units = 32, activation = "relu", input_shape = c(25)) %>% 
    layer_dense(units = 16, activation = "relu") %>% 
    layer_dense(units = 8, activation = "relu") %>% 
    layer_dense(units = 1, activation = 'exponential', name = 'middle', 
    weights = list(rep(0,8) %>% matrix(nrow = 8, ncol = 1),
    array(poiss_hom)))  

  middle_mod = keras_model(middle_in, middle_layers)
  
  main_output = main %>% time_distributed(middle_mod)
  main_output_N <- layer_multiply(list(main_output,inputs[[1]])) %>% layer_flatten()
  
  ### apply main model to ICES
  # Apply time_distributed and create combined_loss_layer
  outputs <- list()
  losses <- list()
  
  combined_loss_params <- list(
    DrivAge = list(lambda_smooth = 100, lambda_mono = 0, direction = 1),
    VehAge = list(lambda_smooth = 10, lambda_mono = 0, direction = -1),
    BonMal = list(lambda_smooth = 10,  lambda_mono = 100,  direction = 1),
    Density = list(lambda_smooth = 10,  lambda_mono = 100, direction = 1),
    VehPower = list(lambda_smooth = 10, lambda_mono = 100, direction = 1)
  )

  
  for (name in names(ice_embed_vectors)) {
    outputs[[name]] <- ice_embed_vectors[[name]] %>% time_distributed(middle_mod)
    args = combined_loss_params[[gsub("_ice_embed", "", name)]]
    smooth_layer = CombinedLossLayer(lambda_smooth = args[[1]], 
                                     lambda_smooth2 = args[[2]],
                                     lambda_mono=args[[3]], 
                                     lambda_fl = args[[4]], 
                                     direction=args[[5]])
    losses[[name]] <- outputs[[name]] %>% layer_flatten() %>% smooth_layer()
  }
  
  losses_concat = losses %>% unname %>% layer_concatenate()
  
  all_out = list(main_output_N, losses_concat) %>% layer_concatenate()
\end{lstlisting}

\clearpage
\pagebreak

\section{Appendix: Local ICEnet plots} \label{local_plots}

\begin{figure}[htb!] 
	\begin{center}
		\begin{minipage}{0.9\textwidth}
			\begin{center}
				\includegraphics[width=\linewidth]{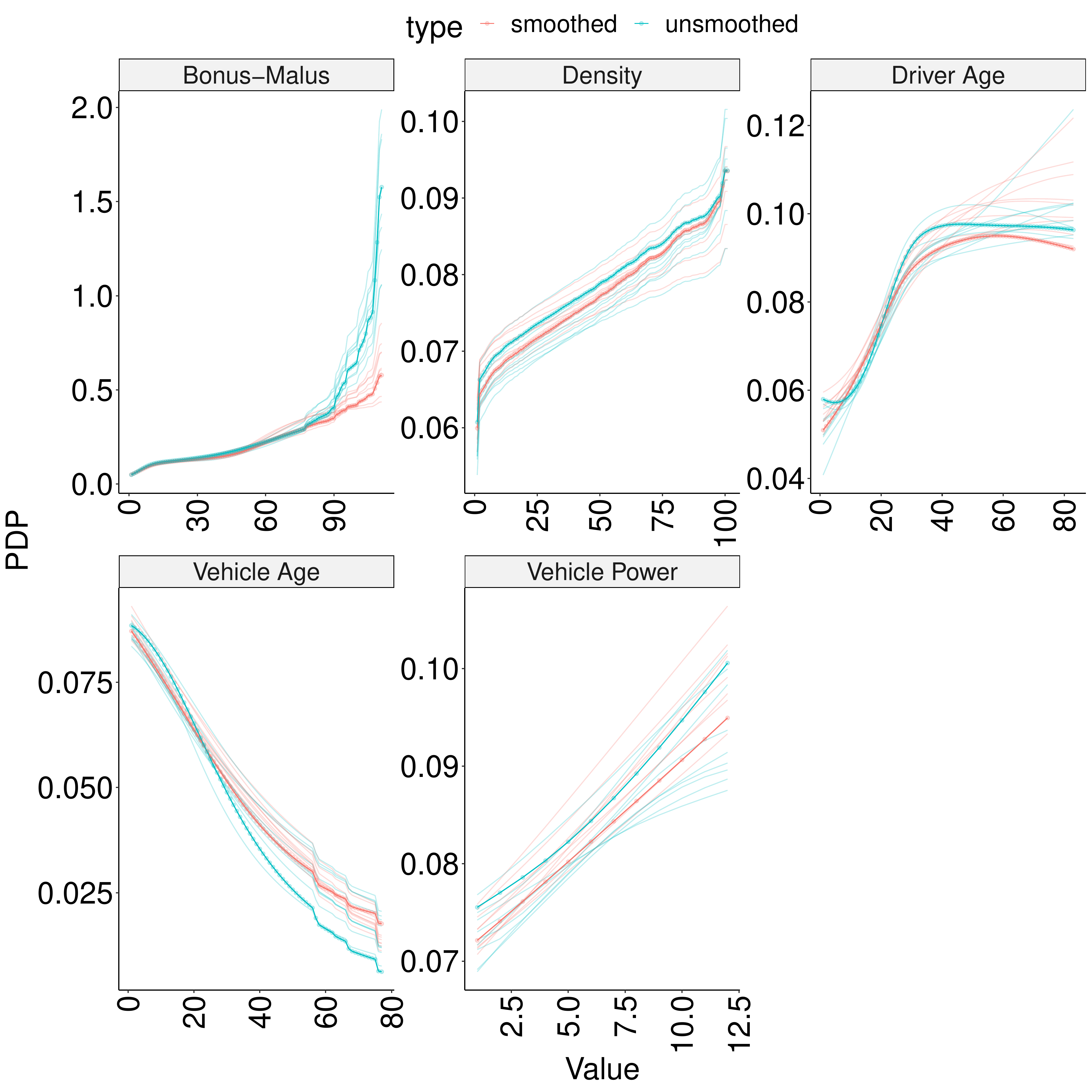}
			\end{center}
		\end{minipage}
	\end{center}
	\caption{PDPs for each of the Bonus-Malus Level, Density, Driver Age, Vehicle Age and Vehicle Power fields shown in separate panels, test set only. Blue lines are PDPs from the FCNs (unsmoothed) and red lines are PDPs from the Local ICEnet (smoothed). Bold lines relate to the PDPs from the first of 10 runs; the lighter lines relate to the remaining runs. Note that the scale of the $y$-axis varies between each panel.}
	\label{pdp_window}
\end{figure}

\begin{figure}[htb!] 
	\begin{center}
		\begin{minipage}{0.9\textwidth}
			\begin{center}
				\includegraphics[width=\linewidth]{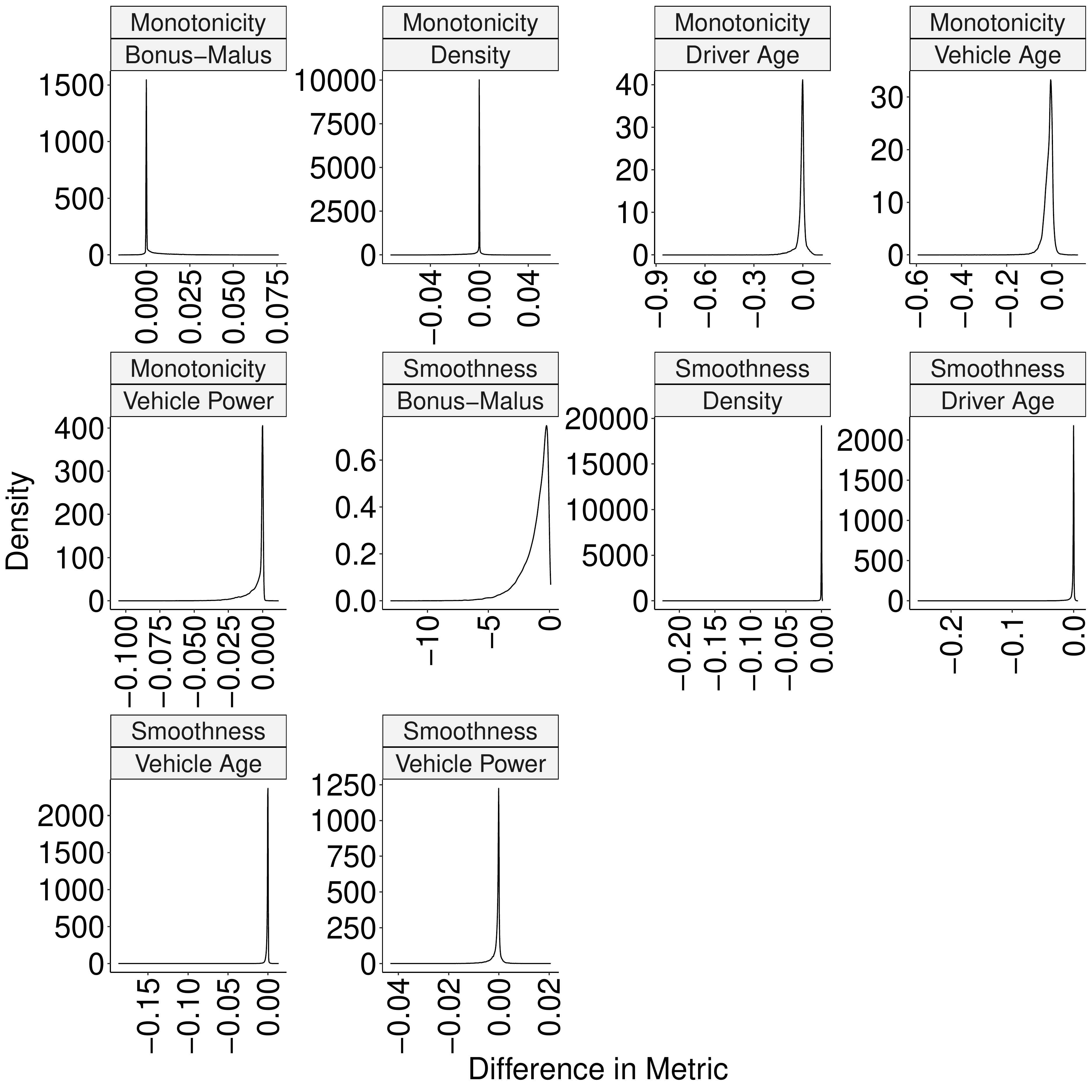}
			\end{center}
		\end{minipage}
	\end{center}
	\caption{Density plots of the difference between the monotonicity and smoothness components of the Local ICEnet loss function \eqref{ICEnet_loss} evaluated for each instance  in the test set.}
	\label{ice_dens_window}
\end{figure}

\begin{figure}[htb!] 
	\begin{center}
		\begin{minipage}{0.9\textwidth}
			\begin{center}
				\includegraphics[width=\linewidth]{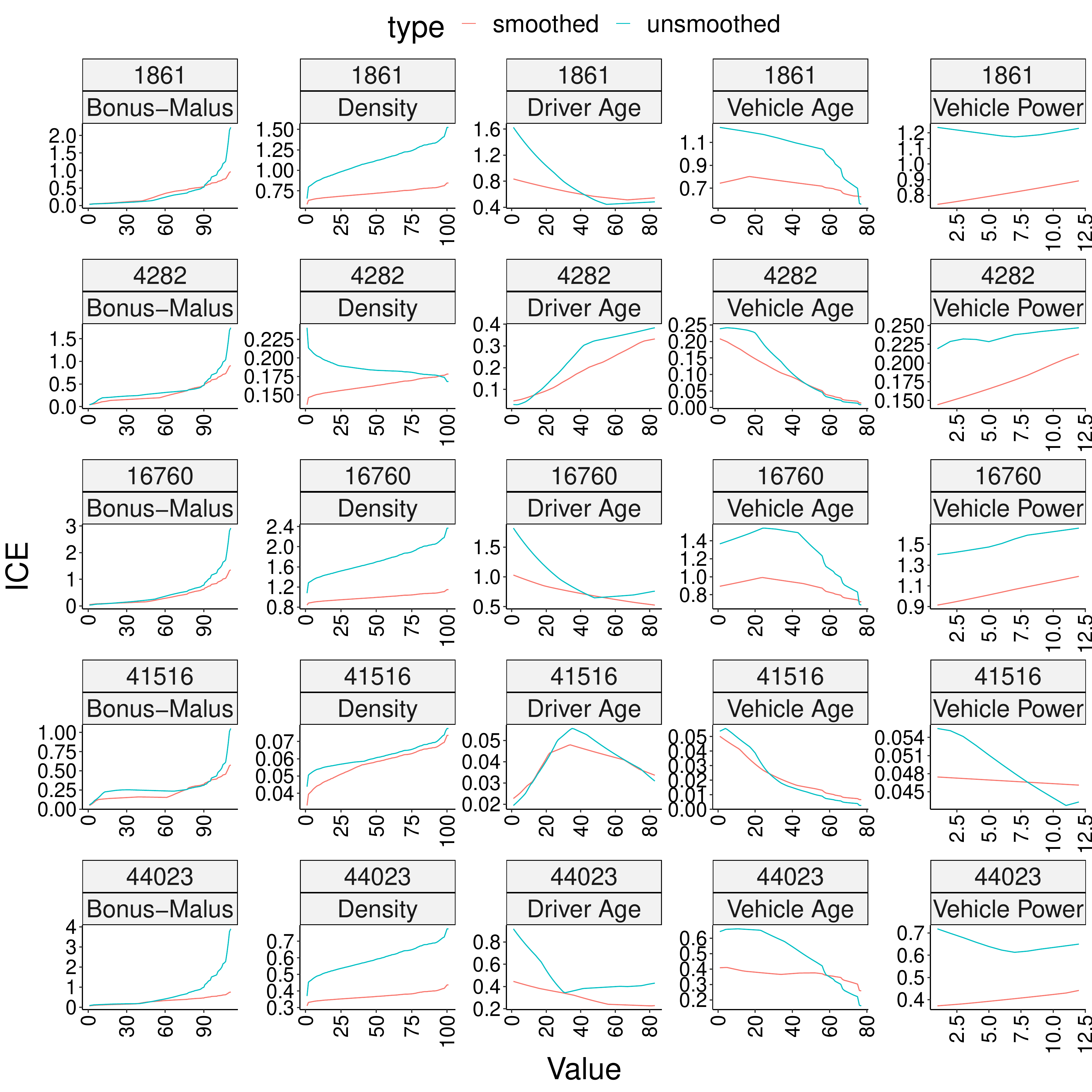}
			\end{center}
		\end{minipage}
	\end{center}
	\caption{ICE plots of the output of the FCN and the Local ICEnet for instances $n$ chosen to be the least monotonic based on the monotonicity score evaluated for each instance in the test set on the outputs of the FCN. Note that the smoothed model is the Local ICEnet and unsmoothed model is the FCN.}
	\label{ices_mon_window}
\end{figure}

\begin{figure}[htb!] 
	\begin{center}
		\begin{minipage}{0.9\textwidth}
			\begin{center}
				\includegraphics[width=\linewidth]{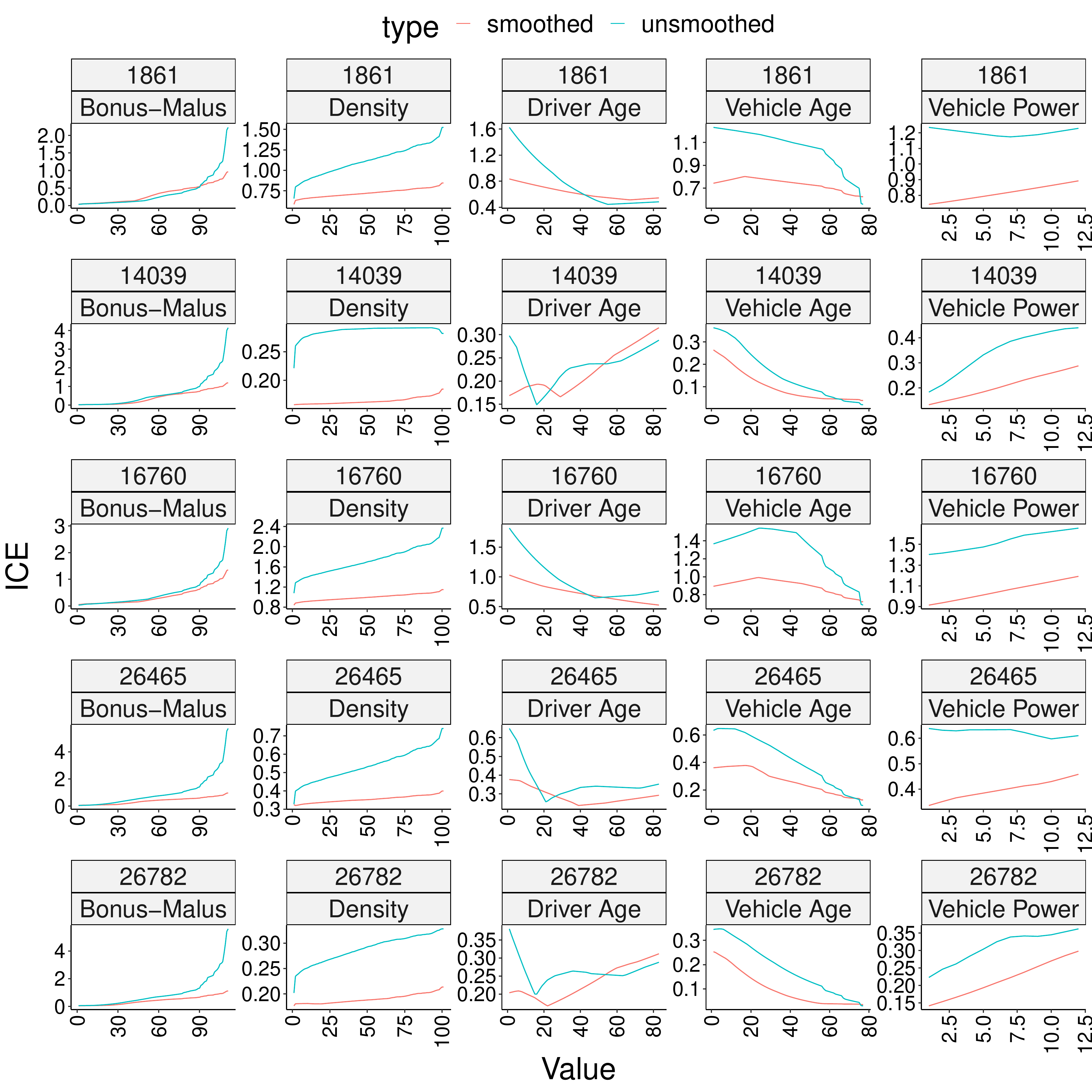}
			\end{center}
		\end{minipage}
	\end{center}
	\caption{ICE plots of the output of the FCN and the Local ICEnet for instances $n$ chosen to be the least smooth based on the smoothness score evaluated for each instance in the test set on the outputs of the FCN. Note that the smoothed model is the Local ICEnet and unsmoothed model is the FCN.}
	\label{ices_smooth_window}
\end{figure}

\end{document}